\documentclass{article}
 
\usepackage{microtype}
\usepackage{graphicx}
\usepackage{booktabs} 

\usepackage{hyperref}

\PassOptionsToPackage{round,authoryear}{natbib}

\usepackage[preprint]{neurips_2026}

\usepackage{amsmath}
\usepackage{amssymb}
\usepackage{amsthm}

\theoremstyle{plain}
\newtheorem{theorem}{Theorem}[section]

\newtheorem{corollary}[theorem]{Corollary}

\theoremstyle{definition}

\newtheorem{proposition}[theorem]{Proposition}
\theoremstyle{remark}

\usepackage{algorithm}
\usepackage{algorithmic}

\newcommand{\RETURN}{\STATE \textbf{return} }

\usepackage{listings}

\usepackage{xurl}

\usepackage{soul}

\usepackage{wrapfig}

\usepackage{tabularx}
\usepackage{multirow}
\usepackage{graphicx} 
\usepackage{enumerate}
\usepackage{enumitem}
\usepackage{tikz}
\usepackage{amsmath}
\usepackage{aliascnt,lscape}
\usepackage{amsfonts}
\usepackage{mdframed}
\usepackage{multicol}
\usepackage{calc}
\usepackage[capitalize,noabbrev]{cleveref}

\usepackage{wrapfig}
\usepackage{float}
\usepackage{thm-restate}
\usepackage{xstring}
\usepackage{pifont}
\usepackage{xspace}
\usepackage{orcidlink}
\usepackage{enumitem}

\usetikzlibrary{arrows.meta}
\tikzset{>={Latex[width=0.5mm,length=1mm]}}

\usetikzlibrary{decorations.pathreplacing}

\newcommand{\qedsymb}{\hfill{\rule{2mm}{2mm}}}
\renewenvironment{proof}{\begin{trivlist} \item[\hspace{\labelsep}{\bf
\noindent Proof.\/}] }{\qedsymb\end{trivlist}}

	\makeatletter
	\newcommand\parag{%
   \@startsection{paragraph}{4}{\z@}%
       {-3\p@ \@plus -1\p@ \@minus -1\p@}%
       {-0.2em \@plus -0.12em \@minus -0.05em}%
       {\normalfont\normalsize\scshape}}

	\makeatother

\newcommand{\MyParagraph}[1]{\medskip \noindent {\bf #1}}
\renewcommand{\paragraph}{\MyParagraph}

\definecolor{yellow}{rgb}{0.8, 0.7, 0.0}
\definecolor{brown}{rgb}{0.59, 0.29, 0.0}
\definecolor{green}{rgb}{0.0, 0.5, 0.0}
\definecolor{orange}{rgb}{1.0, 0.55, 0.0}

\definecolor{highlight-gray}{gray}{0.90}
\newcommand{\gamechange}[2][highlight-gray]{{\setlength{\fboxsep}{0pt}\colorbox{#1}{\ifmmode$\displaystyle#2$\else#2\fi}}}

\newcommand{\node}{\mathsf{root}}

\newcommand{\pre}{\mathsf{\pre}\mbox{-}}

\usepackage{pifont}
\newcommand{\cmark}{\ding{51}}%
\newcommand{\xmark}{\ding{55}}%

\newif\ifdraftcomments
\draftcommentsfalse  

\ifdraftcomments
    \newcommand{\mingxun}[1]{{\footnotesize\color{blue}[Mingxun: #1]}}
    \newcommand{\gf}[1]{{\footnotesize\color{magenta}[GF: #1]}}
    \newcommand{\sara}[1]{{\footnotesize\color{green}[Sara: #1]}}
\else
    \newcommand{\mingxun}[1]{}
    \newcommand{\gf}[1]{}
    \newcommand{\sara}[1]{}
\fi
\newcommand{\oldcomment}[1]{}

\newcounter{cnt:challenge}

\newcommand{\ignore}[1]{}

\newcommand{\name}{\textsc{MaxShapley}\xspace}

\begin{document}

\title{\name: Towards Incentive-compatible Generative Search with Fair Context Attribution}

\author{%
  Sara Patel$^{*\dagger}$ \quad Mingxun Zhou$^{*\mathsection\dagger}$ \quad Giulia Fanti$^*$ \\[0.5em]
  $^*$Carnegie Mellon University \qquad $^\mathsection$HKUST \\
}

\maketitle

\begin{abstract}
    Generative search engines based on large language models (LLMs) are replacing traditional search, fundamentally changing how information providers are compensated. To sustain this ecosystem, we need fair mechanisms to attribute and compensate content providers based on their contributions to generated answers. We introduce MaxShapley, an efficient algorithm for fair credit attribution in generative search pipelines that retrieve external sources before generation. MaxShapley is a special case of the celebrated Shapley value; it leverages a decomposable max-sum utility function to compute attributions with polynomial-time computation in the number of documents, as opposed to the exponential cost of Shapley values. We evaluate MaxShapley on three multi-hop QA datasets (HotPotQA, MuSiQUE, MS MARCO); MaxShapley achieves comparable attribution quality to exact Shapley computation, while consuming a fraction of its tokens---for instance, it gives up to a 9x reduction in resource consumption over prior state-of-the-art methods at the same attribution accuracy. We release open-source code and recalibrated datasets. An educational demo is available at \url{https://fair-search.com}.

\end{abstract}

\section{Introduction}

Generative search engines based on large language models are rapidly replacing traditional search \citep{gholami2026beyond}.
Generative search engines like Perplexity AI~\citep{perplexity} and Google Gemini~\citep{googlegemini} often follow a retrieval-augmented generation (RAG) pattern~\citep{lewis2021retrievalaugmentedgenerationknowledgeintensivenlp,guu2020realmretrievalaugmentedlanguagemodel,izacard2022atlasfewshotlearningretrieval}: they first retrieve relevant documents from a large corpus, then generate a concise response shown directly to the user.

Despite its convenience, generative search \emph{fundamentally disrupts incentive structures for content providers}.
Traditional search directed users to source websites, where providers monetized visits through advertising~\citep{zeff1999advertising}.
Generative search instead answers directly in the interface, reducing visits to original sources.
Traffic losses are already visible for many content providers~\citep{pew2025,thecurrent2025}: Seer Interactive estimates that, as of September 2025, click-through rates to organic content are 62\% lower when Google's AI Overviews are present than when they are not~\citep{seer}.
According to recent reports~\citep{economist-ai-report,similarweb}, the fraction of worldwide web traffic produced by traditional search fell about 5\% from June 2024 to June 2025, with some sources estimating a drop of up to 25\%~\citep{bain2}.
Some media organizations call this decline an ``extinction-level event''~\citep{npr}.
At the same time, source citations in generative search, typically produced through prompt engineering, have been shown to be highly unreliable~\citep{strauss2025attribution} and easily manipulated~\citep{nestaas2025adversarial}.

Content providers are starting to push back: several lawsuits have already been filed against generative search providers for reduced traffic and lost revenue~\citep{chegg,penske,helena}.
A complementary set of lawsuits targets AI companies' use of copyrighted material during training, including the New York Times lawsuit against OpenAI~\citep{nyt_v_openai_2023} and the LibGen lawsuit against Anthropic~\citep{libgenvsanthropic}.
These lawsuits are resulting in billions of dollars in liabilities and growing distrust from content creators~\citep{anthropicbillions2025}.

Industry responses are emerging: some generative search engines promise provider compensation~\citep{gist,oreilly}, and infrastructure providers now let sites block AI crawlers or charge per crawl~\citep{cloudflare}.
Yet the compensation rules remain opaque, with no clear evidence that compensation is tied to a source's \emph{relevance} to the generated answer.

\paragraph{Problem statement and status quo.}
Long-term, the business model for generative search engines will need to evolve to compensate content providers for their contributions~\citep{khosrowi2024engaging}.
Early academic efforts to rethink the LLM ads ecosystem have primarily focused on sponsored search auctions for LLMs~\citep{perplexity-ads,bergemann2024data,hajiaghayi2024ad,duetting2024mechanism,banchio2024ads,feizi2023online}, which do not directly benefit organic content providers.
In this paper, we tackle the algorithmic \textit{attribution} piece of this puzzle: our goal is to define a method for attributing generative search results to specific original sources, so that content providers can be fairly compensated for their work.
In particular, we define fairness according to common axiomatic properties (\Cref{sec:prelim}).
A key operational requirement is that our algorithm should be practical for existing generative search pipelines by minimizing the number and size of queries to an LLM oracle.

\begin{figure}[t]
	\centering
	\begin{minipage}[t]{0.48\linewidth}
		\centering
		\textbf{\small (a) Attribution quality vs.\ token cost}\\[2pt]
		\includegraphics[width=\linewidth]{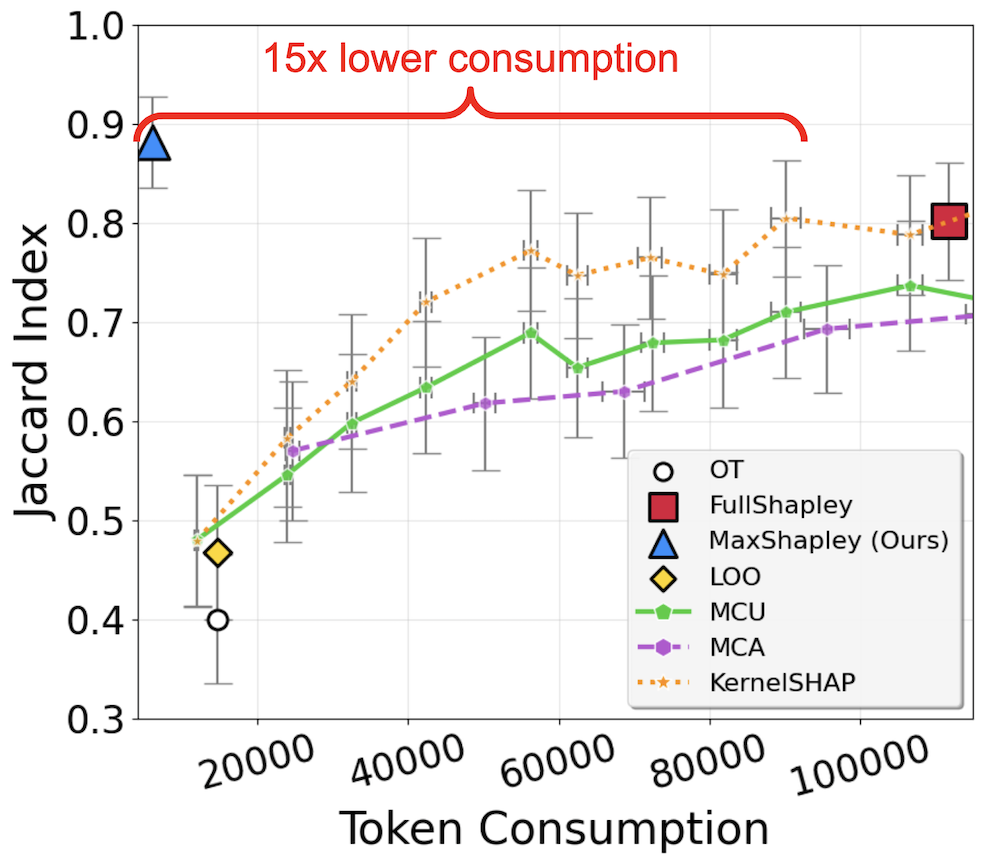}
	\end{minipage}
	~~
	\hfill
	\begin{minipage}[t]{0.48\linewidth}
		\centering
		\textbf{\small (b) \name setting}\\[2pt]
		\includegraphics[width=0.9\textwidth]{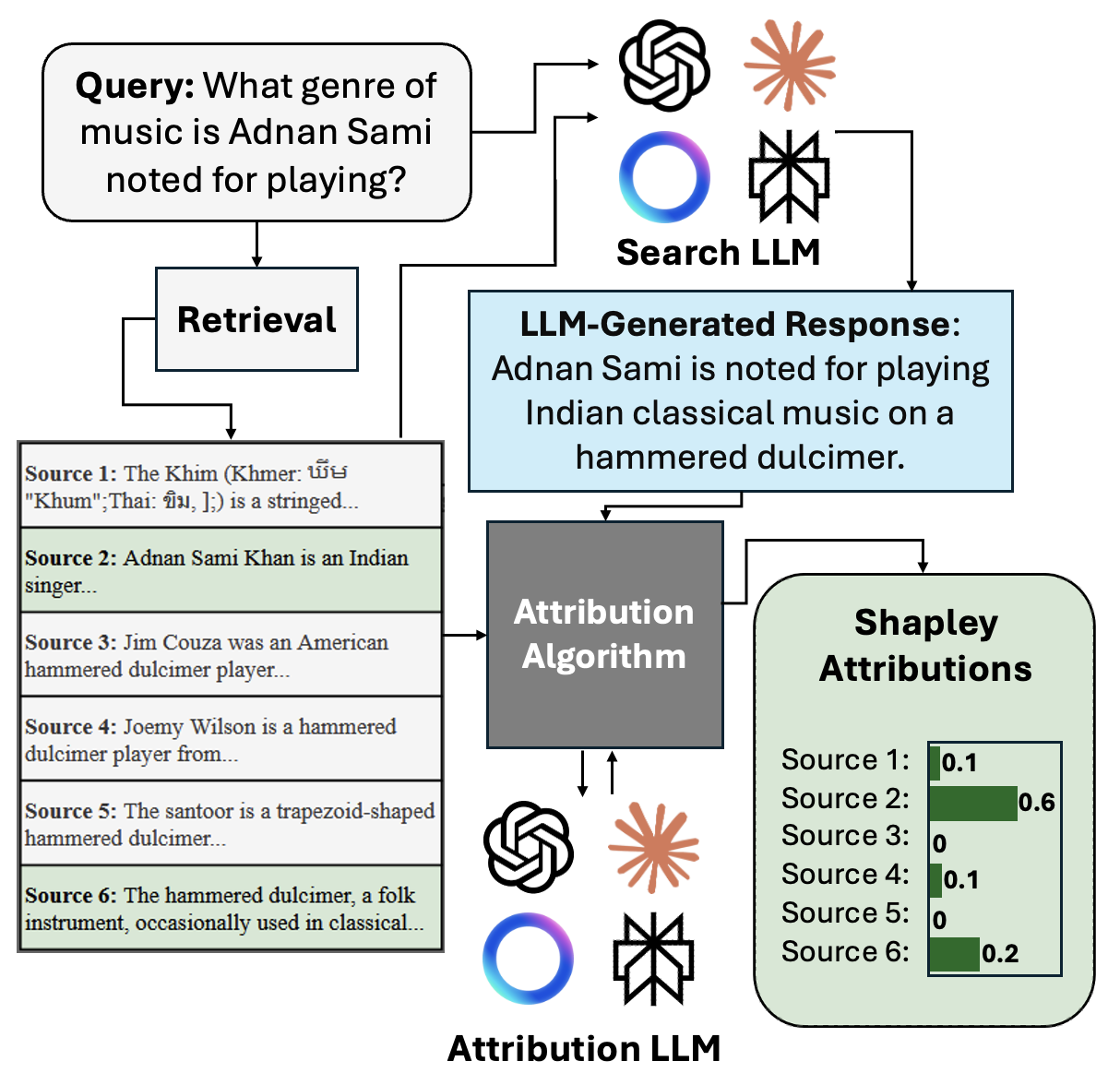}
	\end{minipage}
	\caption{\textbf{Left:} On MuSiQUE with GPT-4.1 nano, \name reaches Jaccard 0.88 ($\uparrow$) using $5.4\%$ of FullShapley's tokens; KernelSHAP needs $15\times$ \name's tokens for similar quality. \textbf{Right:} \name acts in a black-box setting, parallel to a search LLM. It takes as input a query $q$, an answer $a$, and source documents $S$, and outputs attribution scores for each document.
	}
	\label{fig:1}
	\label{fig:setup}
	\vspace{-10pt}
\end{figure}

In the broader machine learning community, variants of the attribution problem have been used to interpret and explain the behaviors of complex machine learning models, which is crucial for building trustworthy and reliable AI systems.
Notable high-impact works include Datamodels~\citep{datamodel}, TRAK~\citep{trak} and Data Shapley~\citep{datashapley2019,datashapley2025,wang2024economic} for training-time attribution to training samples, and LIME~\citep{lime} and KernelSHAP~\citep{kernelshap} for inference-time attribution between inputs and features.
In contrast, our work addresses inference-time credit attribution from generated answers in generative search to retrieved information sources.

Another relevant line of work is \textit{context attribution for retrieval-augmented LLMs}, which aims to identify which retrieved context spans support or explain the final answer generated by an LLM
~\citep{contextcite, learning-to-attribute,qi2024model,selfcite,ding2024attention,liu2025attribot,tokenshapley, hirsch2025laquer}.
However, most existing context attribution methods focus on \textit{fine-grained explainability}, debugging, or citation, rather than \textit{fair and quantitative} credit attribution to information providers from an economic perspective.
A few recent works are closer to our setting and explore the use of Shapley value~\citep{shapley1951notes}
for attribution to information sources~\citep{tokenshapley, nematov2025source, tracllm}, where \citet{nematov2025source} found that the KernelSHAP method~\citep{kernelshap} outperforms other Shapley-based baselines in terms of both attribution accuracy and computational efficiency.
However, Shapley-based attribution typically requires repetitive trials to evaluate the contributions of sources, a known limitation of Shapley values~\citep{nematov2025source,tracllm}, making prior work computationally infeasible in latency-sensitive search settings (\Cref{fig:1}, \Cref{sec:evaluation}).

%

\subsection{Our Contribution}

We study fair attribution of information sources for  incentive-compatible generative search.
\ignore{

\paragraph{Main Insight.}
The incentive allocation mechanism of traditional search engines happens through a competition among the information providers for user traffic, either through search engine optimization (SEO) techniques to improve their ranking in the search results, or auction-based advertisements to directly attract user clicks.
This structure inherently encourages the equilibrium to form that balances the information providers' costs and the revenue from user traffic.

On the contrary, the incentive allocation mechanism of generative search engines is fundamentally different as users typically stop at the LLM-generated answers without visiting the original sources.
The incentive allocation should happen through a more direct credit attribution mechanism initiated by the generative search providers, which should be made fair-by-design to incentivize high-quality information providers' participation.
To this end, there must be a fair information-value attribution mechanism to measure the contribution of each information provider to the final answer, which can then be used to allocate credit accordingly.
}
We propose \name, a novel Shapley-value-based attribution algorithm that quantifies each source's contribution to the final answer by treating providers as players in a cooperative game.
Compared to prior methods~\citep{kernelshap,contextcite, nematov2025source}, \name offers two key advantages:

\name operates downstream of the generative search pipeline: it only takes the query, retrieved sources, and generated response as input (see Figure~\ref{fig:setup}, right). This design has two practical advantages. First, \textbf{\name requires only black-box access}: it does not need token-probability or log-likelihood access to the search LLM, which is required by ContextCite-style and log-likelihood-based RAG attribution utilities~\citep{contextcite,nematov2025source}. It also does not assume reference answers and remains customizable to different evaluation criteria, such as relevance, accuracy, and helpfulness. Second, \textbf{\name is computationally efficient because it avoids repeated context ablations}. Existing utility-based context attribution methods remove or sample subsets of input sources, and then evaluate the impact of the ablation to derive the marginal contribution of each source~\citep{contextcite,kernelshap,nematov2025source}. This ablation loop is the dominant cost because each sampled subset requires another model or utility evaluation. In contrast, \name scores source-keypoint support once and computes the Shapley values of our novel \textit{decomposable max-sum utility function} exactly. After scoring, the attribution step is deterministic arithmetic with no subset sampling; our sorted-scan algorithm computes the exact Shapley values for each max-function subproblem in $O(m\log m)$ time, improving over the $O(m^2)$ routine of prior work~\citep{kernelshap}.



Empirically, \name preserves attribution quality while substantially reducing token cost. It achieves strong correlation with brute-force Shapley values computed from an aggregate LLM-as-a-judge utility function (Kendall-$\tau > 0.5$) and strong alignment with human annotations (Jaccard index $> 0.85$), using less than $6\%$ of the token cost (Figure~\ref{fig:sixpanel}). Compared to KernelSHAP, our strongest Shapley approximation baseline, \textbf{\name reaches the same alignment with human annotations (Jaccard) at less than $7\%$ the cost} (Figure~\ref{fig:1}	, Figure~\ref{fig:sixpanel}).
Our contributions are:


\begin{enumerate}[leftmargin=*]
	\item \textbf{Algorithmic:} We propose \name (\Cref{sec:construction}), a novel and efficient Shapley-value-based algorithm for fair attribution of information sources in generative search, including an exact $O(m\log m)$ algorithm for the max-function Shapley subproblem underlying our construction.
	We show that \name can be viewed as a special case of a more general class of coverage games, which present path towards corroboration-aware attribution (\Cref{app:coverage-games}).
	\item \textbf{Empirical:} We demonstrate through extensive evaluations (\Cref{sec:evaluation}) that \name achieves a significantly better tradeoff between attribution quality and efficiency than existing methods. These gains are robust to choices of attribution LLM and other hyperparameters. 
	\item \textbf{Artifacts:} We release open-source code, manually re-annotated subsets of HotPotQA, MuSiQUE, and TREC\footnote{\url{https://github.com/spaddle-boat/MaxShapley}}, and a demonstration\footnote{\url{https://fair-search.com}} of \name.
\end{enumerate}


%
%
%

\section{Problem Setup and Preliminaries}
\label{sec:prelim}


We study fair source-credit attribution for generative search systems.
For a user query $q$, the system retrieves an ordered list of sources
$S=(s_1,\ldots,s_m)$, where each source is a document, passage, or snippet.
The search LLM $\Psi$ then generates the answer
\[
    a=\Psi(q,S).
\]
The attribution task is to assign a score $\phi_i$ to each source $s_i$, quantifying its contribution to the displayed answer $a$.
These scores can be used for source ranking, auditing, or credit allocation.

We use index sets to denote source sets. 
Let $[m]=\{1,\ldots,m\}$.
For any subset $I\subseteq [m]$, write $S_I=\{s_i:i\in I\}$ for the corresponding subset of sources.
The attribution module has access to the query $q$, the answer $a$, and black-box access to an attribution LLM $\Psi_A$, which may be different from the search LLM $\Psi$.
In evaluation only, some datasets provide a ground truth reference answer $\tilde a$ (distinct from the LLM response $a$)
and binary relevance labels $\tilde y_i\in\{0,1\}$ indicating whether source $s_i$ is annotated as relevant. \name does not require labels or reference answers.

\paragraph{Cooperative game and Shapley attribution.}
A cooperative game consists of a set of players and a utility function that assigns a value to every subset of players.
In our attribution problem, the players are the retrieved sources, indexed by $[m]$.
We represent a game by a subset utility
\[
    U:2^{[m]}\to \mathbb{R}.
\]
The value $U(I)$ represents the utility obtained when only the sources in $I$ are available.
We shift utilities so that $U(\emptyset)=0$; subtracting a constant from all subset values does not change any marginal contribution.
An attribution rule maps each utility function $U$ to a vector
$\boldsymbol{\phi}^U=(\phi_1^U,\ldots,\phi_m^U)$.
We use the standard Shapley desiderata~\citep{shapley1953value}:
\begin{enumerate}[leftmargin=*]
	\item \textbf{Efficiency}: $\sum_{i=1}^m \phi_i^U = U([m])$.
    \item \textbf{Symmetry}: if $U(I\cup\{i\})=U(I\cup\{j\})$ for every $I\subseteq [m]\setminus\{i,j\}$, then $\phi_i^U=\phi_j^U$.
    \item \textbf{Null player}: if $U(I\cup\{i\})=U(I)$ for every $I\subseteq [m]\setminus\{i\}$, then $\phi_i^U=0$.
    \item \textbf{Additivity}: for utilities $U_1,U_2$, the attribution for $U_1+U_2$ satisfies $\phi_i^{U_1+U_2}=\phi_i^{U_1}+\phi_i^{U_2}$.
\end{enumerate}
The Shapley value is the unique attribution rule satisfying these four axioms.
Equivalently, it is the expected marginal contribution of a source when sources arrive in a uniformly random order:
\begin{align}
\phi_i^U
    = \mathbb{E}_{\pi\sim \mathrm{Unif}(\Pi_m)}
    \left[
        U(P_i^\pi\cup\{i\})-U(P_i^\pi)
    \right],
\label{eq:shapley-prelim}
\end{align}
where $\Pi_m$ is the set of permutations of $[m]$, and $P_i^\pi$ is the set of indices appearing before $i$ in permutation $\pi$.
Shapley value is notorious for its computational intractability:
without special structure in the utility $U$,
exact Shapley computation requires enumeration over all $2^m$ subsets.

\subsection{The cooperative game in generative search and baseline attribution methods}

Generative search can be viewed naturally from a cooperative game perspective: the retrieved sources jointly provide necessary context for the LLM to generate the final answers.
To formally define the game, the most straightforward choice is a black-box answer-quality utility function defined by re-running the full search-generation pipeline on each subset $I \subseteq [m]$ of retrieved sources and measure the quality of the answer.
\begin{align}
    a_I
    &= \Psi(q,S_I)
    && \text{\emph{answer generation from a source subset}} \\
    \widetilde U(I)
    &= \text{\textsc{Judge}}_{\Psi_A}(q,a_I,S_I;p)
    && \text{\emph{answer-quality grading}} \\
    U(I)
    &= \widetilde U(I)-\widetilde U(\emptyset)
    && \text{\emph{marginal gain over the null case}}
    \label{eq:black-box-utility}
\end{align}
Here $p$ denotes the judging prompt and scoring rule, which may encode criteria such as correctness, relevance, completeness, and, for oracle baselines, optional access to a reference answer.
Following the LLM-as-a-judge paradigm~\citep{zheng2023judgingllmasajudgemtbenchchatbot,liu2023gevalnlgevaluationusing}, $\text{\textsc{Judge}}_{\Psi_A}$ is implemented by prompting the attribution LLM $\Psi_A$ to score the generated answer under these criteria (prompt in Appendix~\ref{app:prompts}). 

\paragraph{FullShapley.}
This baseline computes the exact Shapley value in Equation~\ref{eq:shapley-prelim} for the black-box utility in Equation~\ref{eq:black-box-utility}.
It is the cleanest fairness baseline, but it is exponentially expensive in the number of retrieved sources.
In our evaluation, FullShapley is also given access to the reference answer when available, making it a strong oracle-style baseline.

\paragraph{Leave-One-Out (\textbf{LOO}) and Optimal Transport (\textbf{OT}).}
LOO attribution~\citep{influencefunction, liu2025attribot} assigns
\[
    \phi_i^{\mathsf{LOO}} = U([m])-U([m]\setminus\{i\}).
\]
It is cheap and intuitive, but not budget balanced: the scores need not sum to $U([m])$.
It also under-credits redundant support; if two sources provide the same evidence, removing either one may barely change the utility.
We additionally evaluate a sentence-level LOO variant in Appendix~\ref{app:sentence_loo}.
For OT, we adapt GiLOT~\citep{li2024gilot} by treating each source document as the perturbation unit and measuring influence through optimal-transport distances between generations with(out) that source.

\paragraph{Monte-Carlo Shapley (\textbf{MCU}, \textbf{MCA}).}
Monte-Carlo methods approximate Equation~\ref{eq:shapley-prelim} by sampling permutations and averaging observed marginal contributions~\citep{michalak2013efficient,mitchell2022sampling}.
\textbf{MCU} samples permutations uniformly.
\textbf{MCA} pairs each sampled permutation with its reverse to reduce variance.
These methods approach the Shapley value with enough samples, but in LLM settings each sampled subset still corresponds to an expensive utility evaluation.

\paragraph{KernelSHAP.}
{KernelSHAP}~\citep{kernelshap} estimates Shapley values by fitting a weighted linear surrogate model over sampled subsets.
It is a strong general-purpose approximation, but its quality depends on the number and distribution of sampled subsets; empirically, it requires far more LLM calls than \name to approach FullShapley-quality attribution (\Cref{sec:evaluation}).

\section{\name: Efficient and Fair Attribution in Generative Search}
\label{sec:construction}

We present \name, a Shapley-based algorithm for fair attribution in generative search.
\begin{wrapfigure}[16]{r}{0.5\textwidth}
		\centering
		\resizebox{\linewidth}{!}{\definecolor{sysblue}{RGB}{36,94,151}
\definecolor{sysbluefill}{RGB}{241,247,253}
\definecolor{sysgreen}{RGB}{52,128,77}
\definecolor{sysgreenfill}{RGB}{240,249,243}
\definecolor{syspurple}{RGB}{103,70,157}
\definecolor{syspurplefill}{RGB}{247,244,252}
\definecolor{sysorange}{RGB}{221,118,32}
\definecolor{sysorangefill}{RGB}{255,247,239}
\definecolor{sysgray}{RGB}{83,88,97}

\begin{tikzpicture}[
    x=1cm,y=1cm,
    every node/.style={font=\scriptsize},
    outer/.style={rounded corners=5pt, line width=0.55pt, draw=sysblue!70, fill=sysbluefill},
    block/.style={rounded corners=4pt, line width=0.55pt, fill=white, align=center, inner sep=3pt},
    arrow/.style={-{Latex[length=1.8mm,width=1.05mm]}, line width=0.68pt, draw=sysgray},
    best/.style={fill=sysgreen!18, draw=sysgreen!85, line width=0.55pt},
    cell/.style={draw=sysgray!42, line width=0.35pt, minimum width=0.43cm, minimum height=0.27cm, inner sep=0pt, align=center},
    bar/.style={minimum height=0.16cm, anchor=west, inner sep=0pt}
]

\draw[outer] (0,0) rectangle (6.85,4.84);
\node[anchor=west, font=\bfseries\scriptsize, text=sysblue!70!black,
      text width=6.35cm, align=left] at (0.18,4.58)
{Input: Retrieved Sources $s_1,\ldots,s_m$\\[-1pt]
and RAG Generated Answer $a$};

\node[block, draw=sysblue!75, text width=2.65cm, minimum height=1.82cm] (decomp) at (1.72,3.34) {};
\node[font=\bfseries\scriptsize, text=sysblue!75!black] at (1.72,3.98) {Keypoint Decomposition};
\node[align=center] at (1.72,3.60) {Answer $a$};
\node[font=\scriptsize] at (1.72,3.36) {$\Downarrow$};
\node[align=center] at (1.72,2.89) {$n$ keypoints and weights\\[-1pt] $p_1,\dots,p_n$\\[-1pt] $w_1,\dots,w_n$};

\node[block, draw=sysgreen!75, fill=sysgreenfill, text width=2.65cm, minimum height=1.82cm] (scores) at (5.15,3.34) {};
\node[font=\bfseries\scriptsize, text=sysgreen!65!black] at (5.15,4.03) {Support Scores $v_{ij}$};
\node[font=\tiny, text=sysgray, align=center] at (5.15,3.55) {\textit{how well does source $s_i$}\\[-1pt] \textit{support keypoint $p_j$?}};
\node[font=\scriptsize] at (4.76,2.55) {$s_1$};
\node[font=\scriptsize] at (5.15,2.55) {$s_2$};
\node[font=\scriptsize] at (5.54,2.55) {$s_3$};
\node[font=\scriptsize] at (4.35,3.11) {$p_1$};
\node[font=\scriptsize] at (4.35,2.82) {$p_2$};
\node[cell,best] at (4.76,3.11) {1.0};
\node[cell] at (5.15,3.11) {.6};
\node[cell] at (5.54,3.11) {0};
\node[cell] at (4.76,2.82) {0};
\node[cell] at (5.15,2.82) {.2};
\node[cell,best] at (5.54,2.82) {1.0};

\node[block, draw=syspurple!75, fill=syspurplefill, text width=2.65cm, minimum height=1.82cm] (utility) at (1.72,1.02) {};
\node[font=\bfseries\scriptsize, text=syspurple!75!black, align=center] at (1.72,1.56) {The Sum-Max\\[-1pt] Cooperative Game};
\node[font=\scriptsize] at (1.72,1.07) {Subset $I$ has utility $U(I)$};
\node[font=\scriptsize, align=center] at (1.72,0.62) {$=\sum_{j=1}^n w_j\cdot$\\[-1pt] $\left(\max_{i\in I} v_{i,j}\right)$};

\node[block, draw=sysorange!85, fill=sysorangefill, text width=2.65cm, minimum height=1.82cm] (credit) at (5.15,1.02) {};
\node[font=\bfseries\scriptsize, text=sysorange!75!black] at (5.15,1.66) {Shapley Attribution};
\node[anchor=east] at (4.52,1.22) {$s_1$};
\node[bar, fill=sysorange, minimum width=0.70cm] at (4.57,1.22) {};
\node[anchor=west, font=\tiny] at (5.31,1.22) {.350};
\node[anchor=east] at (4.52,0.94) {$s_2$};
\node[bar, fill=sysblue, minimum width=0.40cm] at (4.57,0.94) {};
\node[anchor=west, font=\tiny] at (5.01,0.94) {.200};
\node[anchor=east] at (4.52,0.66) {$s_3$};
\node[bar, fill=sysgreen, minimum width=0.90cm] at (4.57,0.66) {};
\node[anchor=west, font=\tiny] at (5.51,0.66) {.450};

\draw[arrow] (decomp.east) -- (scores.west);
\draw[arrow] (scores.south) .. controls (5.15,2.27) and (1.72,2.27) .. (utility.north);
\draw[arrow] (utility.east) -- (credit.west);

\end{tikzpicture}}
	\caption{\name pipeline: answer decomposition, source-keypoint support scoring, max-sum utility, and exact Shapley attribution. \label{fig:maxshapley-overview}}
\end{wrapfigure}
Our main design is a novel utility function for generative search and an efficient algorithm for its Shapley values.
Figure \ref{fig:maxshapley-overview} illustrates the pipeline.
We observe that retrieved information sources in generative search provide both complementary and overlapping information, and attribution should account for both:

\begin{itemize}[leftmargin=*]
    \item \textbf{\underline{Cooperation} in providing \underline{complementary} information.} Sources contribute distinct information to the final answer (e.g., articles covering different market sectors for a stock trend query), forming a \textit{cooperative} game.

    \item \textbf{\underline{Competition} in providing \underline{overlapping} information.} When sources cover overlapping perspectives of the same information, the more relevant source should receive more credit, showing a \textit{competitive} nature even in the cooperative game.
\end{itemize}

We define the utility in two steps. First, the attribution LLM $\Psi_A$ decomposes the answer $a$ into key points $P=\{p_1,\dots,p_n\}$ (top left of Figure \ref{fig:maxshapley-overview}; prompt in Appendix \ref{app:prompts}); examples are in Appendix \ref{app:keypoint_examples}. Second, for each source-keypoint pair, $\Psi_A$ assigns a support score $v_{i,j}$ measuring how well source $s_i$ supports key point $p_j$ (top right of Figure \ref{fig:maxshapley-overview}; prompt in Appendix \ref{app:prompts}). The simple pairwise interface uses $nm$ LLM calls; batched prompts can reduce API requests to $m$ calls (one source against all key points) or $n$ calls (one key point against all sources), with comparable token/FLOP cost up to batching and KV-cache constants. After that, all utility and Shapley computations are deterministic.

For a subset $I\subseteq [m]$, each key point is covered by its best supporting source in $S_I$, and the subset utility is the weighted sum of these best-support scores:
\begin{align}
U_{\name}(I) = \sum_{j=1}^n w_j\max_{i\in I} v_{i,j},
\label{eq:utility-def}
\end{align}
where $\max_{i\in\emptyset}v_{i,j}=0$ (bottom left, Figure \ref{fig:maxshapley-overview}).
Here, $w_j$ is the weight of key point $p_j$.
Given the support scores $v_{ij}$, evaluating $U_{\name}$ requires no further LLM calls.

This sum-max structure resembles the MaxSim score in ColBERT~\citep{colbert, colbertv2}, a state-of-the-art retrieval method~\citep{colbertv2}, which further justifies our design. Our approach differs in two ways: (1) we compute scores at the \emph{key-point} level via LLM-as-a-judge rather than at the token level, capturing holistic semantic information; (2) we connect this structure to fair attribution through Shapley values (\Cref{sec:shapley-computation}). In fact, we also conducted experiments with an attribution mechanism based on Shapley values with the original MaxSim score, but the results showed that the token-level embeddings fail to distinguish between low-level linguistic similarity and deeper-level of semantic relevance.

\subsection{Efficient Shapley value computation for the new utility function}
\label{sec:shapley-computation}

\begin{algorithm}[t]
\caption{\name attribution algorithm}
\label{alg:full-algorithm}
\begingroup
\small
\renewcommand{\algorithmicrequire}{\textsc{Input:}}
\renewcommand{\algorithmicensure}{\textsc{Output:}}
\renewcommand{\algorithmiccomment}[1]{\texttt{//} \emph{#1}}
\newcommand{\algstage}[1]{\STATE \textit{\textbf{#1}}}
\begin{algorithmic}
\REQUIRE Query $q$, retrieved sources $S=\{s_1,\dots,s_m\}$, generated answer $a$, attribution LLM $\Psi_A$.
\ENSURE Attribution score $\phi_i$ for each source $s_i$.
\algstage{Stage I: Generating support scores}
\STATE Use $\Psi_A$ to decompose answer $a$ into key points $P=\{p_1,\dots,p_n\}$ and weights $w_1,\dots,w_n$.
\FOR{$j=1$ to $n$}
    \FOR{$i=1$ to $m$}
        \STATE Use $\Psi_A$ to score how well source $s_i$ supports key point $p_j$; denote the score by $v_{i,j}$.
    \ENDFOR
\ENDFOR
\algstage{Stage II: Computing Shapley values for the utility function $U_{\name}$}
\STATE Initialize $\phi_i\leftarrow 0$ for every source $i\in[m]$.
\FOR{$j=1$ to $n$}
    \STATE $\{\phi^{\mathsf{Max}}_{i,j}\}_{i=1}^m \leftarrow \textsc{MaxGameShapley}(v_{1,j},\dots,v_{m,j})$.
    \FOR{$i=1$ to $m$}
        \STATE $\phi_i\leftarrow \phi_i+w_j\phi^{\mathsf{Max}}_{i,j}$.
    \ENDFOR
\ENDFOR
\RETURN $\phi_1,\dots,\phi_m$.
\STATE \rule{0.96\linewidth}{0.4pt}
\algstage{Subroutine \textsc{MaxGameShapley}$(v_1,\dots,v_m)$: Shapley values for $U(I)=\max_{i\in I}v_i$}
\STATE Sort players so that $0\le v_1\le v_2\le \cdots \le v_m$, preserving original indices.
\STATE Set $v_0\leftarrow 0$ and $c\leftarrow 0$.
\FOR{$r=1$ to $m$}
    \STATE $c\leftarrow c + (v_r-v_{r-1})/(m-r+1)$ \COMMENT{distribute the marginal value to players $\{r,r+1,\dots,m\}$}
    \STATE $\phi_r\leftarrow c$
\ENDFOR
\RETURN $\{\phi_i\}_{i\in[m]}$, mapped back to the original source order.
\end{algorithmic}
\endgroup
\end{algorithm}

A key advantage of this utility function is that it admits exact Shapley value computation in polynomial time, avoiding Monte Carlo approximations entirely.
First, note that the sum-max structure of Equation~\ref{eq:utility-def} decomposes into $n$ independent maximization games, one per key point.
Define
\begin{align}
	U_{\textsc{Max}}^{j}(I) = \max_{i\in I} v_{i,j},
	\label{eq:per-keypt-utility}
\end{align}
with corresponding Shapley values $\phi^{U_{\textsc{Max}}^{j}}_{i}$, for $j\in [n]$.
The additivity axiom of Shapley values implies that
$\phi^{U_{\name}}_{i} = \sum_{j=1}^n w_j \phi^{U_{\textsc{Max}}^{j}}_{i}$.
Hence, we only need the per-keypoint Shapley values $\phi^{U_{\textsc{Max}}^{j}}_{i}$ based on the utility in \eqref{eq:per-keypt-utility}.

\paragraph{Shapley value for key-point level maximization games.}
Each per-keypoint game $U_{\textsf{Max}}^{j}(I) = \max_{i\in I} v_{i,j}$ is a maximization game, a special class with a particularly simple exact Shapley computation.
By Corollary~\ref{cor:max-game-shapley}, its exact Shapley values can be computed by sorting the $m$ scores once and scanning the interval endpoints. The \textsc{MaxGameShapley} subroutine in Algorithm~\ref{alg:full-algorithm} implements this corollary in $O(m\log m)$ time, compared to $O(m \cdot 2^m)$ for brute-force Shapley and the $O(m^2)$ Max SHAP routine described by \citet{kernelshap}.
The intuition is geometric: source $i$ covers the interval $[0,v_i]$ on a one-dimensional support axis, and each interval segment is shared equally by the sources that cover it. \Cref{app:coverage-games} formalizes this as a special case of a general information coverage game.
Thus, after the source-keypoint scores are fixed, no subset enumeration or quadratic pairwise scan is needed.

The full \name is presented in Algorithm~\ref{alg:full-algorithm}, with implementation details in Appendix~\ref{app:implementation}.

\begin{proposition}
	\label{prop:maxshapley-complexity}
	Given source-keypoint support scores $\{v_{i,j}\}$ and key-point weights $\{w_j\}$, \name (Algorithm~\ref{alg:full-algorithm}) returns the exact Shapley values $\phi_i^{U_{\name}}$ for the best-support utility $U_{\name}$ in Equation~\ref{eq:utility-def}.
	Its computational complexity is $O((T+nmL)|\Psi|+nm\log m)$, where $T$ upper bounds the token budget of the non-repeated \name prompts and generated outputs, $L$ upper bounds each source-keypoint scoring prompt length, including both the prompt template and the source length, and $|\Psi|$ is the maximum per-token FLOP cost of the attribution LLM. 
\end{proposition}
The proof follows from Shapley additivity and the max-game computation in Appendix~\ref{app:max-game-coverage}.

\paragraph{Comparison to Related Attribution Methods.}
\label{sec:comparison-to-related-attribution-methods}
Table~\ref{tab:theoretical_comparison} compares access assumptions, Shapley desiderata, and attribution-time inference cost under a normalized per-token compute model.

\begin{table}[t]
	\centering
	\scriptsize
	\setlength{\tabcolsep}{1.6pt}
	\renewcommand{\arraystretch}{1.02}
	\begin{tabularx}{\textwidth}{p{1.65cm}p{2.3cm}ccccX}
	\toprule
	\textbf{Method} & \textbf{Access} & \multicolumn{4}{c}{\textbf{Shapley desiderata}} & \textbf{Attribution-time cost} \\
	\cmidrule(lr){3-6}
	& & \textbf{Eff.} & \textbf{Sym.} & \textbf{Null} & \textbf{Add.} & \\
	\midrule
	\multicolumn{7}{l}{\emph{Black-box methods}} \\
	FullShapley & Black box & \cmark & \cmark & \cmark & \cmark & $O(2^m(mL+T)|\Psi| + m2^m)$ \\
	LOO & Black box & \xmark & \cmark & \cmark & \cmark & $O(m(mL+T)|\Psi| + m)$ \\
	MCU/MCA & Black box & \cmark$^*$ & \cmark$^*$ & \cmark$^*$ & \cmark$^*$ & $O(km(mL+T)|\Psi| + km)$ \\
	\textbf{\name (Ours)} & \textbf{Black box} & \cmark & \cmark & \cmark & \cmark & $\mathbf{O((T+nmL)|\Psi| + nm\log m)}$ \\
	\midrule
	\multicolumn{7}{l}{\emph{Model-internal / training-based methods}} \\
	AttriBot~\citep{liu2025attribot} & Model/proxy & \xmark & \cmark & \cmark & \cmark & Target LOO: $O(m(mL+T)|\Psi|)$; proxy replaces $|\Psi|$ with $|\Psi_{\mathrm{proxy}}|$ \\
	ARC-JSD~\citep{li2025attributing} & Logits & \xmark & \cmark & \cmark & \cmark & $O(m(mL+T)|\Psi| + mT|\mathcal V|)$ \\
	SelfCite~\citep{selfcite} & Logprobs/train$^\dagger$ & N/A & N/A & N/A & N/A & Per statement: $O(N(mL+T)|\Psi|)$ plus optional SimPO training \\
	\bottomrule
	\end{tabularx}
    \caption[Comparison of attribution methods by access, desiderata, and compute]{Comparison of attribution methods. $m$: sources; $n$: key points; $L$: source-level scoring context length (including the source length); $T$: generated-output budget; $k$: sampled permutations; $N$: SelfCite best-of-$N$ candidates; $|\mathcal V|$: vocabulary size; $|\Psi|$: per-token model cost. $^*$Satisfied in expectation. $^\dagger$Requires training.}
	\label{tab:theoretical_comparison}
  \end{table}

\ignore{

\subsection{Generalized Construction of \name}

\mingxun{TODO}

In the most generalized form,
we consider the decomposition of the answer generation process as a logical induction graph, where each node represents a piece of information (e.g., a retrieved document, an intermediate reasoning step, or the final answer), and each directed edge represents a logical dependency between two pieces of information (e.g., a reasoning step that derives one piece of information from another).

}

\section{Empirical Evaluation}
\label{sec:evaluation}

\begin{figure*}[htbp]
    \centering
    \begin{minipage}[t]{0.32\textwidth}
     \centering
     \textbf{HotPotQA}
     \includegraphics[width=\textwidth]{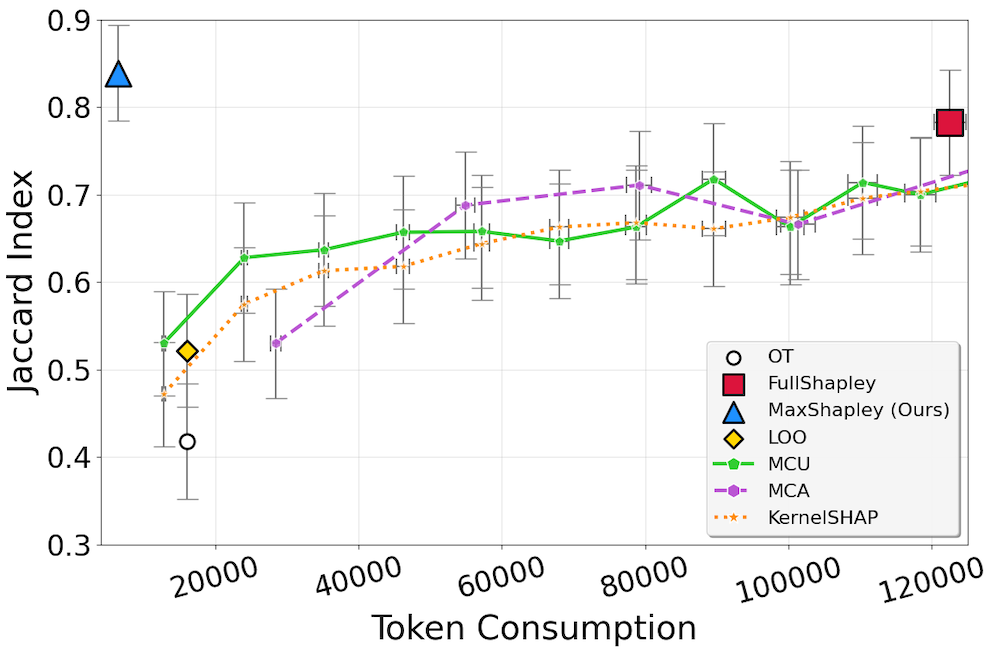}
     \vspace{0.1em}
     \includegraphics[width=\textwidth]{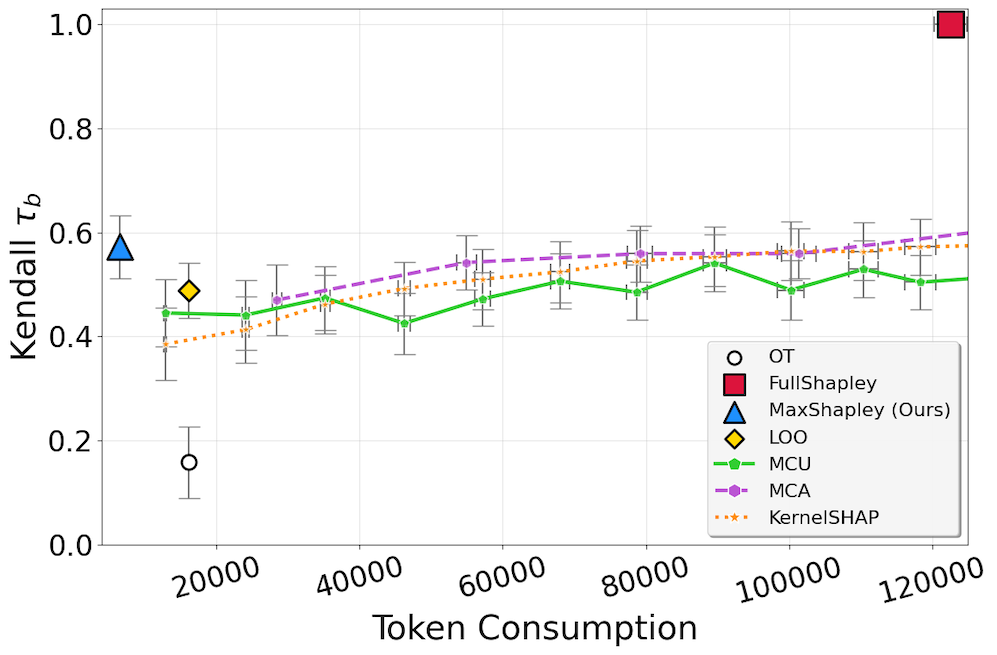}
    \end{minipage}%
    \hfill
    \begin{minipage}[t]{0.32\textwidth}
     \centering
     \textbf{MS-MARCO}
     \includegraphics[width=\textwidth]{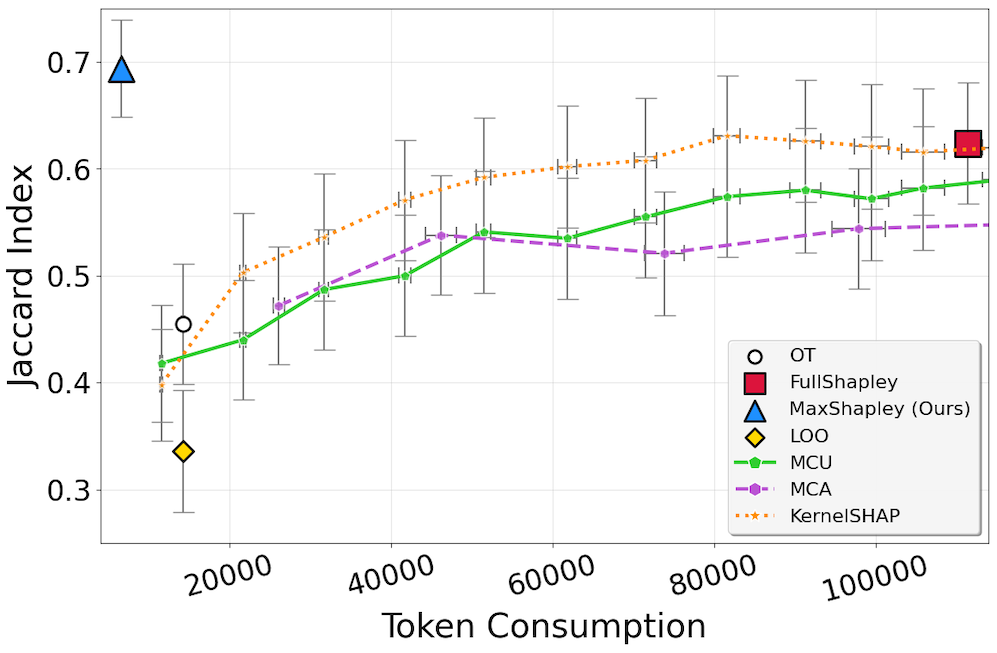}
     \vspace{0.1em}
     \includegraphics[width=\textwidth]{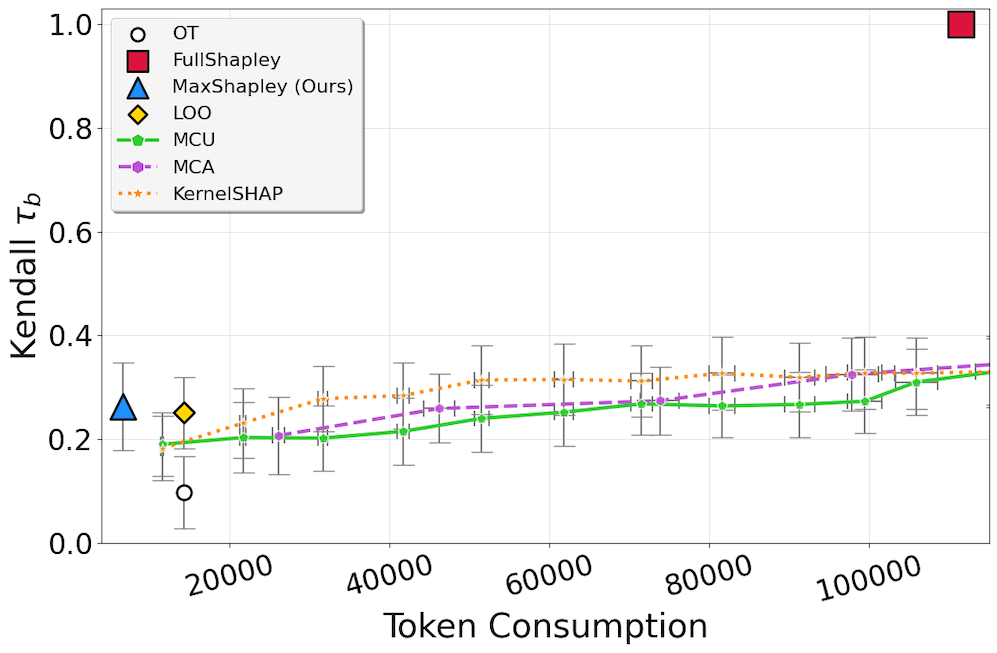}
    \end{minipage}%
    \hfill
    \begin{minipage}[t]{0.32\textwidth}
     \centering
     \textbf{MuSiQUE}
     \includegraphics[width=\textwidth]{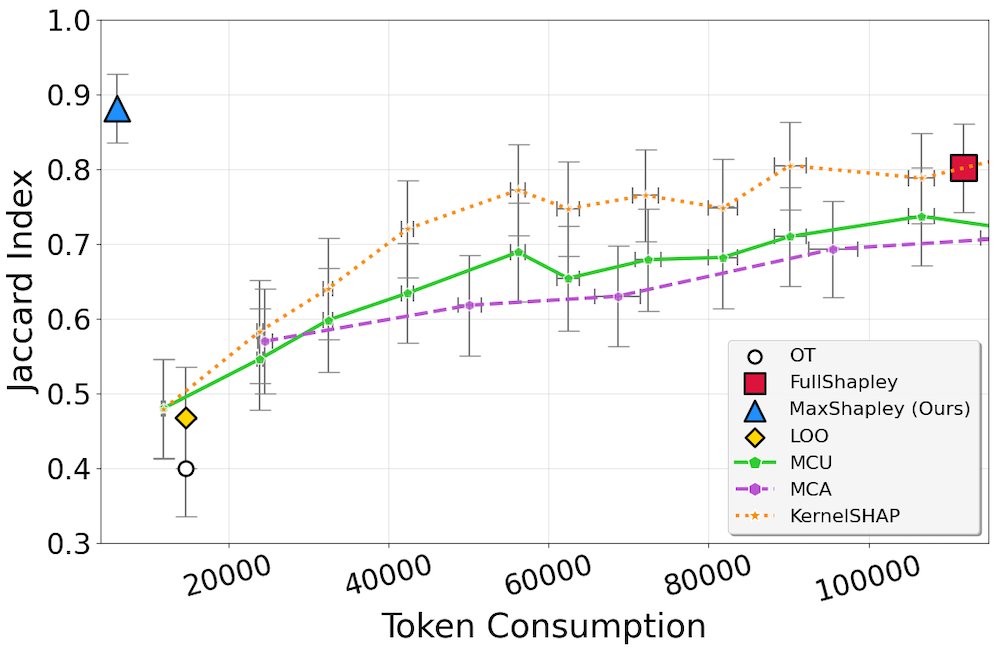}
     \vspace{0.1em}
     \includegraphics[width=\textwidth]{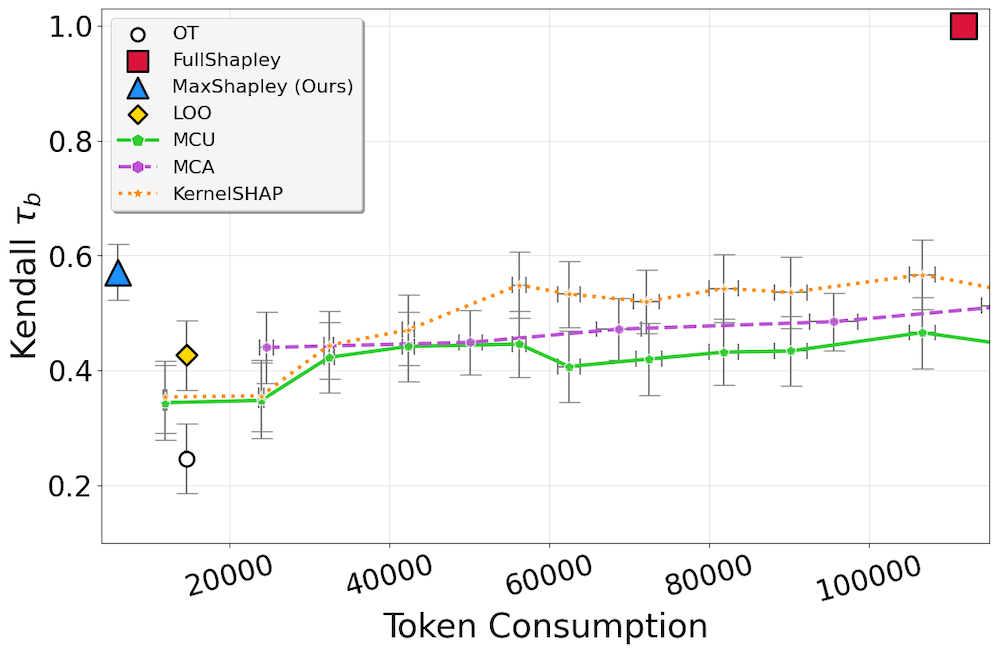}
    \end{minipage}
    \caption[]{Quality of attribution (Jaccard index w.r.t. ground truth (top), Kendall $\tau_b$ w.r.t. FullShapley (bottom)) versus token consumption for attribution algorithms on all three datasets, using GPT-4.1 nano. MaxShapley and KernelSHAP achieve comparable Jaccard index with FullShapley with the latter
    using at least 12$\times$ more tokens.
    MaxShapley reaches a strong ordinal correlation via Kendall's $\tau_b$ with FullShapley for HotPotQA and MuSiQUE. On MS-MARCO, MaxShapley reaches a moderate ordinal correlation. For similar correlations with FullShapley, KernelSHAP consumes 4-9$\times$ more tokens than MaxShapley.}
    \label{fig:sixpanel}
   \end{figure*}

\subsection{Evaluation Setup}

We evaluate \name in terms of
\textit{quality of attribution} and \textit{efficiency} of the algorithm.
In this evaluation, we set \name's key-point weights uniformly, $w_j=1/n$, to avoid adding another LLM-scored component to the pipeline.


\paragraph{Baselines.}
We compare to the baselines introduced in \Cref{sec:prelim},
including \textbf{FullShapley}, \textbf{LOO}, \textbf{OT}, \textbf{MCU}, \textbf{MCA}, and \textbf{KernelSHAP}. Since we view FullShapley as our strongest baseline, we gave its $\textbf{Judge}$ function access to a ground truth response $\tilde a$ (Prompt in \Cref{app:prompts}).



\paragraph{Quality Metrics.}
These metrics measure the agreement with \emph{FullShapley} and/or groundtruth labels:
\begin{itemize}[leftmargin=*]
	  \item \textbf{Jaccard@\(K\)} between the ground truth relevance labels for each document (see Datasets below) and the top-$K$ elements of the \name vector. Let $R$ be the ground truth relevant sources for a query (as annotated in a dataset) and let $K=|R|$. Let $T=\text{Top}_K(\hat{\boldsymbol{\phi}})$.
	$\mathrm{Jaccard@}K=\frac{|T\cap R|}{|T\cup R|}$. Jaccard@K $\in [0, 1]$ with 1.0 indicating perfect agreement between sets (higher is better).
  \item \textbf{Kendall’s $\tau_b$~\citep{kendall1938new}} between the \name and FullShapley vectors. Ordinal agreement between rankings induced by $\hat{\boldsymbol{\phi}}$ and $\boldsymbol{\phi}^\star$; $\tau_b\in[-1,1]$ with 0.0 indicating no ordinal correlation and 1.0 indicating perfect correlation (higher is better).
\end{itemize}

\paragraph{Datasets.}
We evaluate three multi-hop question answering datasets:
\textbf{(1) HotpotQA}~\citep{yang2018hotpotqadatasetdiverseexplainable}: Reasoning queries over Wikipedia documents.
\textbf{(2) MuSiQUE}~\citep{trivedi2022musiquemultihopquestionssinglehop}: Structured two-hop questions in full-Wiki setting.
\textbf{(3) MS MARCO}~\citep{bajaj2018msmarcohumangenerated, craswell2020overviewtrec2019deep}: Passage ranking with graded relevance judgments from the TREC 2019 Deep Learning Track.

\paragraph{Annotation}
Because the original dataset labels can be noisy, we manually re-annotated focused subsets of 100 HotPotQA queries, 100 MuSiQUE queries, and 95 MS-MARCO queries, with six candidate sources per query and consensus labels from two annotators.
Annotation protocol and agreement statistics are in Appendix~\ref{app:annotation}.
We include an ablation varying the number of sources in Appendix \ref{app:scaling_exp} to empirically compare scaling behavior between \name and KernelSHAP.
All methods are evaluated on these annotated subsets across 100 independent runs; we report means and standard errors. We also evaluate \name on the original, larger datasets in Appendix~\ref{app:largeexp}.

\paragraph{Evaluation Limitations.}
We treat both FullShapley and manually-annotated relevance as ground truth for attribution quality, although neither is perfect. LLM-as-a-judge exhibits scoring inconsistencies even at temperature 0 (\Cref{sec:results}), affecting all Shapley methods, including FullShapley. Additionally, manually-annotated relevance measures a related but distinct concept from Shapley attribution. 
Lacking a single ground truth, we measure association with both of these quantities.

\subsection{Main Results}
\label{sec:results}

\paragraph{(1) \name achieves a better tradeoff among baselines between attribution quality efficiency by a significant margin.}
Figures \ref{fig:sixpanel} show the tradeoff between token consumption and attribution quality, measured by Jaccard index with ground truth and Kendall's $\tau_b$ with FullShapley, using GPT-4.1 nano. \name consistently outperforms LOO, OT, MCU, MCA, and KernelSHAP across all datasets and metrics. On MuSiQUE, KernelSHAP at least 15$\times$ more tokens than \name to achieve comparable Jaccard index scores, while Monte Carlo methods require 15-20$\times$ more tokens.

For rank correlation measured by Kendall's $\tau_b$ (Figure \ref{fig:sixpanel}, bottom), MaxShapley achieves a strong ordinal correlation with FullShapley on MuSiQUE and HotPotQA, while KernelSHAP requires 7-9$\times$ more tokens to reach the same correlation quality. On MS MARCO, MaxShapley achieves a moderate correlation with KernelSHAP achieving the same with 4$\times$ more tokens.\footnote{There is no standard way to interpret the quality of a $\tau_b$ correlation, we follow \citet{wicklin2023interpret}, using $\ge 0.71$ to indicate very strong correlation,
 $>=0.49$ to indicate a strong, $>=0.26$ for moderate correlation, and $<0.26$ for weak/negligible.}

On MS-MARCO, we observe a degradation in the quality of attribution across all Shapley attribution methods.
MS-MARCO, unlike HotPotQA and MuSiQUE, is a less curated dataset, with sometimes confusing information source content (even for humans). As such, the Search LLM had more trouble forming coherent and correct responses to queries with a given set of information sources.

Note that \name has a higher Jaccard index with the ground truth than FullShapley on HotPotQA and MS-MARCO (Figure~\ref{fig:sixpanel}). This is consistent with the sensitivity of whole-answer subset scoring discussed below: semantically similar regenerated answers can receive different \(\text{\textsc{Judge}}_{\Psi_A}\) scores, which affects FullShapley-style baselines.

In addition, we provide the evaluation results on the full, original MuSiQUE dataset in \Cref{app:ablations}.


\paragraph{(2) Localized LLM judging stabilizes Shapley attribution.}
The black-box utility in Equation~\ref{eq:black-box-utility} is a natural baseline: for each source subset, it re-runs generation and asks \(\text{\textsc{Judge}}_{\Psi_A}\) to score the whole answer.
This couples Shapley attribution to repeated ablation judgments over perturbed answers.
As shown in Table~\ref{tab:inconsistency}, two answers that are semantically equivalent to human readers receive mean \(\text{\textsc{Judge}}_{\Psi_A}\) scores of 0.3 and 1.0 over 10 deterministic runs, while identical inputs show little variance.
This pattern matches prior findings that LLM evaluations are sensitive to wording, source ordering, and context length~\citep{salinas2024butterflyeffectalteringprompts,lin2025llmwhispererinconspicuousattack}.

\name instead uses the attribution LLM for localized semantic judgments: decomposing the displayed answer into key points and scoring source-keypoint support.
Given these local scores, all subset utilities and Shapley values are computed deterministically through Equation~\ref{eq:utility-def}.
This keeps the semantic flexibility of LLM judging while removing repeated whole-answer ablation judgments from the Shapley computation.
The strong Jaccard scores in Figure~\ref{fig:sixpanel} suggest that localized judging preserves semantic signal and reduces attribution instability.

\subsection{Ablations}
\label{sec:ablations}
We conducted an ablation study on \name and baselines.
See more details in Appendix \ref{app:ablations}.

\paragraph{Model Selection.}
For our attribution LLM, we evaluated GPT-4.1 nano (OpenAI), Claude Haiku 4.5, Claude 3.5, and Claude Sonnet 4 (Anthropic), and conducted our main experiments using only the first two. Haiku 4.5 achieved notably higher attribution quality than GPT-4.1 nano with moderately increased token consumption and cost, but with a noticeable increase in execution time, excluded Sonnet 4 due to prompt incompatibility and higher cost and Claude Haiku 3.5, which was retired prior to our dataset expansion and replaced by Haiku 4.5 (see Figure~\ref{fig:model_comparison} in Appendix~\ref{app:ablations}).

\paragraph{Effect of Clipping.}
Despite temperature 0, several baselines assign tiny nonzero attribution scores to otherwise irrelevant sources, making their low-ranked ordering unstable.
To remove this negligible mass,
we clip all attributions below 0.05 and renormalize for every baseline \emph{except} \name, whose max operation does not spread small scores across unsupported sources.
We show the effect of clipping on FullShapley in Appendix \ref{app:clipping}.

\paragraph{Positional Bias.}
LLMs can over-attend to information near context boundaries~\citep{liu2023lostmiddlelanguagemodels}, so source order can affect attribution.
On 30 MuSiQUE samples with exactly two relevant sources, Haiku 3.5 gave \name a 0.12 higher Jaccard score when the relevant sources were placed first than when all sources were randomly shuffled (Appendix~\ref{app:annotation} and \ref{app:model_selection}).
We therefore randomly shuffle source order before each LLM call in our evaluation, preventing any source from systematically benefiting from favorable placement.

\section{Conclusion, Limitations, and Future Directions}
\label{sec:conclusion}

We present \name, a black-box Shapley-based algorithm for fair attribution to information sources in generative search. Our novel max-sum utility formulation preserves the Shapley fairness desiderata while admitting an efficient runtime.
Empirically, \name achieves strong agreement with FullShapley and human relevance annotations while using a small fraction of FullShapley's token cost.
Several limitations and assumptions motivate future work. First, we operate under a black-box access assumption, motivated by a desire for third-party attribution. White-box attribution may also be desirable for some LLM providers, opening the design space. 
Another natural extension is a corroboration-aware utility that assigns additional value when multiple sources mutually support the same key point. 

\begin{ack}
This work was supported in part by the National Science
Foundation under grant CCF-2338772, as well as by the Initiative for Cryptocurrencies and Contracts
(IC3) and the CyLab Secure Blockchain Initiative, together with their respective industry sponsors.

\paragraph{Generative AI Use Disclosure.}
The original technical idea, initial algorithm implementation, and first draft of the paper were done by the human authors. LLMs were utilized to develop the evaluation pipeline and polish the paper's presentation.

\end{ack}

\clearpage



\bibliographystyle{plainnat} 
\bibliography{pir,refs}

\begin{thebibliography}{78}
\providecommand{\natexlab}[1]{#1}
\providecommand{\url}[1]{\texttt{#1}}
\expandafter\ifx\csname urlstyle\endcsname\relax
  \providecommand{\doi}[1]{doi: #1}\else
  \providecommand{\doi}{doi: \begingroup \urlstyle{rm}\Url}\fi

\bibitem[Allen and Netwon(2025)]{cloudflare}
Will Allen and Simon Netwon.
\newblock {Introducing pay per crawl: Enabling content owners to charge AI crawlers for access}.
\newblock \url{https://blog.cloudflare.com/introducing-pay-per-crawl/}, 7 2025.
\newblock {The Cloudflare Blog}. [Online; accessed 2025-10-17].

\bibitem[Allyn(2025)]{npr}
Bobby Allyn.
\newblock {Will Google's AI Overviews kill news sites as we know them?}
\newblock \emph{{NPR Technology}}, July 2025.

\bibitem[{Anthropic}(2024)]{anthropic2024haiku}
{Anthropic}.
\newblock {Claude} 3.5 {Haiku}.
\newblock \url{https://www.anthropic.com/claude/haiku}, 2024.
\newblock Accessed: 2025-10-07.

\bibitem[{Anthropic}(2025)]{anthropic2025claude4}
{Anthropic}.
\newblock Introducing {Claude} 4.
\newblock \url{https://www.anthropic.com/news/claude-4}, 2025.
\newblock Accessed: 2025-10-07.

\bibitem[Athena~Chapekis(2025)]{pew2025}
Anna~Lieb Athena~Chapekis.
\newblock Google users are less likely to click on links when an ai summary appears in the results, 2025.
\newblock URL \url{https://www.pewresearch.org/short-reads/2025/07/22/google-users-are-less-likely-to-click-on-links-when-an-ai-summary-appears-in-the-results/}.

\bibitem[Bajaj et~al.(2018)Bajaj, Campos, Craswell, Deng, Gao, Liu, Majumder, McNamara, Mitra, Nguyen, Rosenberg, Song, Stoica, Tiwary, and Wang]{bajaj2018msmarcohumangenerated}
Payal Bajaj, Daniel Campos, Nick Craswell, Li~Deng, Jianfeng Gao, Xiaodong Liu, Rangan Majumder, Andrew McNamara, Bhaskar Mitra, Tri Nguyen, Mir Rosenberg, Xia Song, Alina Stoica, Saurabh Tiwary, and Tong Wang.
\newblock Ms marco: A human generated machine reading comprehension dataset, 2018.
\newblock URL \url{https://arxiv.org/abs/1611.09268}.

\bibitem[Banchio et~al.(2024)Banchio, Mehta, and Perlroth]{banchio2024ads}
Martino Banchio, Aranyak Mehta, and Andres Perlroth.
\newblock Ads in conversations.
\newblock \emph{arXiv preprint arXiv:2403.11022}, 2024.

\bibitem[{Bartz v. Anthropic PBC, No. 69058235}(2024)]{libgenvsanthropic}
{Bartz v. Anthropic PBC, No. 69058235}.
\newblock U.S. District Court, Central District of California, 2024.
\newblock URL \url{https://www.courtlistener.com/docket/69058235/bartz-v-anthropic-pbc/}.

\bibitem[Bergemann et~al.(2024)Bergemann, Bojko, D{\"u}tting, Leme, Xu, and Zuo]{bergemann2024data}
Dirk Bergemann, Marek Bojko, Paul D{\"u}tting, Renato~Paes Leme, Haifeng Xu, and Song Zuo.
\newblock Data-driven mechanism design: Jointly eliciting preferences and information.
\newblock \emph{arXiv preprint arXiv:2412.16132}, 2024.

\bibitem[Chuang et~al.(2025)Chuang, Cohen-Wang, Shen, Wu, Xu, Lin, Glass, Li, and tau Yih]{selfcite}
Yung-Sung Chuang, Benjamin Cohen-Wang, Zejiang Shen, Zhaofeng Wu, Hu~Xu, Xi~Victoria Lin, James~R. Glass, Shang-Wen Li, and Wen tau Yih.
\newblock Selfcite: Self-supervised alignment for context attribution in large language models.
\newblock In \emph{Forty-second International Conference on Machine Learning}, 2025.
\newblock URL \url{https://openreview.net/forum?id=rKi8eyJBoB}.

\bibitem[Cohen-Wang et~al.(2024)Cohen-Wang, Shah, Georgiev, and Madry]{contextcite}
Benjamin Cohen-Wang, Harshay Shah, Kristian Georgiev, and Aleksander Madry.
\newblock Contextcite: Attributing model generation to context.
\newblock \emph{Advances in Neural Information Processing Systems}, 37:\penalty0 95764--95807, 2024.

\bibitem[Cohen-Wang et~al.(2025)Cohen-Wang, Chuang, and Madry]{learning-to-attribute}
Benjamin Cohen-Wang, Yung-Sung Chuang, and Aleksander Madry.
\newblock Learning to attribute with attention, 2025.
\newblock URL \url{https://arxiv.org/abs/2504.13752}.

\bibitem[Craswell et~al.(2020)Craswell, Mitra, Yilmaz, Campos, and Voorhees]{craswell2020overviewtrec2019deep}
Nick Craswell, Bhaskar Mitra, Emine Yilmaz, Daniel Campos, and Ellen~M. Voorhees.
\newblock Overview of the trec 2019 deep learning track, 2020.
\newblock URL \url{https://arxiv.org/abs/2003.07820}.

\bibitem[Criddle(2024)]{perplexity-ads}
Cristina Criddle.
\newblock Perplexity in talks with top brands on ads model as it challenges google.
\newblock \url{https://www.ft.com/content/ecf299f4-e0a9-468b-af06-8a94e5f0b1f4}, 9 2024.
\newblock Online; accessed 2025-10-16.

\bibitem[DeepMind(2023)]{googlegemini}
Google DeepMind.
\newblock Google gemini: A multimodal ai model.
\newblock Blog post / technical announcement, 2023.
\newblock URL \url{https://blog.google/technology/ai/gemini-collection}.

\bibitem[Ding et~al.(2024)Ding, Luo, Cao, and Luo]{ding2024attention}
Qiang Ding, Lvzhou Luo, Yixuan Cao, and Ping Luo.
\newblock Attention with dependency parsing augmentation for fine-grained attribution.
\newblock \emph{arXiv preprint arXiv:2412.11404}, 2024.

\bibitem[Dubey et~al.(2024)Dubey, Feng, Kidambi, Mehta, and Wang]{auction-llm}
Avinava Dubey, Zhe Feng, Rahul Kidambi, Aranyak Mehta, and Di~Wang.
\newblock Auctions with llm summaries.
\newblock In \emph{Proceedings of the 30th ACM SIGKDD Conference on Knowledge Discovery and Data Mining}, KDD '24, page 713–722. Association for Computing Machinery, 2024.
\newblock ISBN 9798400704901.
\newblock \doi{10.1145/3637528.3672022}.
\newblock URL \url{https://doi.org/10.1145/3637528.3672022}.

\bibitem[Duetting et~al.(2024)Duetting, Mirrokni, Paes~Leme, Xu, and Zuo]{duetting2024mechanism}
Paul Duetting, Vahab Mirrokni, Renato Paes~Leme, Haifeng Xu, and Song Zuo.
\newblock Mechanism design for large language models.
\newblock In \emph{Proceedings of the ACM Web Conference 2024}, pages 144--155, 2024.

\bibitem[Economist(2025)]{economist-ai-report}
The Economist.
\newblock Ai is killing the web. can anything save it?, 2025.
\newblock URL \url{https://www.economist.com/business/2025/07/14/ai-is-killing-the-web-can-anything-save-it}.

\bibitem[Feizi et~al.(2023)Feizi, Hajiaghayi, Rezaei, and Shin]{feizi2023online}
Soheil Feizi, MohammadTaghi Hajiaghayi, Keivan Rezaei, and Suho Shin.
\newblock Online advertisements with llms: Opportunities and challenges.
\newblock \emph{arXiv preprint arXiv:2311.07601}, 2023.

\bibitem[Flynn(2025)]{penske}
Kerry Flynn.
\newblock {Penske Media sues Google over AI summaries taking traffic}.
\newblock \emph{Axios}, 9 2025.
\newblock URL \url{https://www.axios.com/2025/09/14/penske-media-sues-google-ai}.
\newblock Online; accessed 2025-10-18.

\bibitem[Gholami et~al.(2026)Gholami, Firullo, Cheyre, and Acquisti]{gholami2026beyond}
Samira Gholami, Cristiana Firullo, Cristobal Cheyre, and Alessandro Acquisti.
\newblock Beyond search: Llm adoption and web traffic concentration.
\newblock \emph{Available at SSRN 6238578}, 2026.

\bibitem[Ghorbani and Zou(2019)]{datashapley2019}
Amirata Ghorbani and James Zou.
\newblock Data shapley: Equitable valuation of data for machine learning.
\newblock In \emph{International conference on machine learning}, pages 2242--2251. PMLR, 2019.

\bibitem[Gist(2026)]{gist}
Gist.
\newblock {Gist: AI monetization solutions}.
\newblock \url{https://gist.ai/}, 2026.
\newblock Online; accessed 2025-10-17.

\bibitem[Grattafiori et~al.(2024)]{grattafiori2024llama3}
Aaron Grattafiori et~al.
\newblock The {Llama 3} herd of models.
\newblock \emph{arXiv preprint arXiv:2407.21783}, 2024.
\newblock URL \url{https://arxiv.org/abs/2407.21783}.

\bibitem[Gu et~al.(2025)Gu, Jiang, Shi, Tan, Zhai, Xu, Li, Shen, Ma, Liu, Wang, Zhang, Wang, Gao, Ni, and Guo]{gu2025surveyllmasajudge}
Jiawei Gu, Xuhui Jiang, Zhichao Shi, Hexiang Tan, Xuehao Zhai, Chengjin Xu, Wei Li, Yinghan Shen, Shengjie Ma, Honghao Liu, Saizhuo Wang, Kun Zhang, Yuanzhuo Wang, Wen Gao, Lionel Ni, and Jian Guo.
\newblock A survey on llm-as-a-judge, 2025.
\newblock URL \url{https://arxiv.org/abs/2411.15594}.

\bibitem[Gunasekara et~al.(2024)Gunasekara, Hsieh, Le, and Baron]{oreilly}
Lucky Gunasekara, Andy Hsieh, Lan Le, and Julie Baron.
\newblock {The New O'Reilly Answers: The R in ``RAG" Stands for ``Royalties"}.
\newblock \url{https://www.oreilly.com/radar/the-new-oreilly-answers-the-r-in-rag-stands-for-royalties/}, 6 2024.
\newblock Online; accessed 2025-10-17.

\bibitem[Guu et~al.(2020)Guu, Lee, Tung, Pasupat, and Chang]{guu2020realmretrievalaugmentedlanguagemodel}
Kelvin Guu, Kenton Lee, Zora Tung, Panupong Pasupat, and Ming-Wei Chang.
\newblock Realm: Retrieval-augmented language model pre-training, 2020.
\newblock URL \url{https://arxiv.org/abs/2002.08909}.

\bibitem[Hajiaghayi et~al.(2024)Hajiaghayi, Lahaie, Rezaei, and Shin]{hajiaghayi2024ad}
MohammadTaghi Hajiaghayi, S{\'e}bastien Lahaie, Keivan Rezaei, and Suho Shin.
\newblock Ad auctions for llms via retrieval augmented generation.
\newblock \emph{Advances in Neural Information Processing Systems}, 37:\penalty0 18445--18480, 2024.

\bibitem[Hirsch et~al.(2025)Hirsch, Slobodkin, Wan, Stengel-Eskin, Bansal, and Dagan]{hirsch2025laquer}
Eran Hirsch, Aviv Slobodkin, David Wan, Elias Stengel-Eskin, Mohit Bansal, and Ido Dagan.
\newblock Laquer: Localized attribution queries in content-grounded generation.
\newblock \emph{arXiv preprint arXiv:2506.01187}, 2025.

\bibitem[Ilyas et~al.(2022)Ilyas, Park, Engstrom, Leclerc, and Madry]{datamodel}
Andrew Ilyas, Sung~Min Park, Logan Engstrom, Guillaume Leclerc, and Aleksander Madry.
\newblock Datamodels: Understanding predictions with data and data with predictions.
\newblock In Kamalika Chaudhuri, Stefanie Jegelka, Le~Song, Csaba Szepesvari, Gang Niu, and Sivan Sabato, editors, \emph{Proceedings of the 39th International Conference on Machine Learning}, volume 162 of \emph{Proceedings of Machine Learning Research}, pages 9525--9587. PMLR, 17--23 Jul 2022.
\newblock URL \url{https://proceedings.mlr.press/v162/ilyas22a.html}.

\bibitem[Izacard et~al.(2022)Izacard, Lewis, Lomeli, Hosseini, Petroni, Schick, Dwivedi-Yu, Joulin, Riedel, and Grave]{izacard2022atlasfewshotlearningretrieval}
Gautier Izacard, Patrick Lewis, Maria Lomeli, Lucas Hosseini, Fabio Petroni, Timo Schick, Jane Dwivedi-Yu, Armand Joulin, Sebastian Riedel, and Edouard Grave.
\newblock Atlas: Few-shot learning with retrieval augmented language models, 2022.
\newblock URL \url{https://arxiv.org/abs/2208.03299}.

\bibitem[Kendall(1938)]{kendall1938new}
M.~G. Kendall.
\newblock A new measure of rank correlation.
\newblock \emph{Biometrika}, 30\penalty0 (1/2):\penalty0 81--93, 1938.
\newblock ISSN 00063444.
\newblock URL \url{http://www.jstor.org/stable/2332226}.

\bibitem[Khattab and Zaharia(2020)]{colbert}
Omar Khattab and Matei Zaharia.
\newblock Colbert: Efficient and effective passage search via contextualized late interaction over bert.
\newblock In \emph{Proceedings of the 43rd International ACM SIGIR conference on research and development in Information Retrieval}, pages 39--48, 2020.

\bibitem[Khosrowi et~al.(2024)Khosrowi, Finn, and Clark]{khosrowi2024engaging}
Donal Khosrowi, Finola Finn, and Elinor Clark.
\newblock Engaging the many-hands problem of generative-ai outputs: A framework for attributing credit.
\newblock \emph{AI and Ethics}, 2024.

\bibitem[Koh and Liang(2017)]{influencefunction}
Pang~Wei Koh and Percy Liang.
\newblock Understanding black-box predictions via influence functions.
\newblock In \emph{International conference on machine learning}, pages 1885--1894. PMLR, 2017.

\bibitem[Lee et~al.(2025)Lee, Kwon, and Jin]{lee2025gradegeneratingmultihopqa}
Jeongsoo Lee, Daeyong Kwon, and Kyohoon Jin.
\newblock Grade: Generating multi-hop qa and fine-grained difficulty matrix for rag evaluation, 2025.
\newblock URL \url{https://arxiv.org/abs/2508.16994}.

\bibitem[Lewis et~al.(2021)Lewis, Perez, Piktus, Petroni, Karpukhin, Goyal, Küttler, Lewis, tau Yih, Rocktäschel, Riedel, and Kiela]{lewis2021retrievalaugmentedgenerationknowledgeintensivenlp}
Patrick Lewis, Ethan Perez, Aleksandra Piktus, Fabio Petroni, Vladimir Karpukhin, Naman Goyal, Heinrich Küttler, Mike Lewis, Wen tau Yih, Tim Rocktäschel, Sebastian Riedel, and Douwe Kiela.
\newblock Retrieval-augmented generation for knowledge-intensive nlp tasks, 2021.
\newblock URL \url{https://arxiv.org/abs/2005.11401}.

\bibitem[Li et~al.(2025)Li, Chen, Hu, Gao, Wang, and Yilmaz]{li2025attributing}
Ruizhe Li, Chen Chen, Yuchen Hu, Yanjun Gao, Xi~Wang, and Emine Yilmaz.
\newblock Attributing response to context: A jensen-shannon divergence driven mechanistic study of context attribution in retrieval-augmented generation.
\newblock \emph{arXiv preprint arXiv:2505.16415}, 2025.

\bibitem[Li et~al.(2024)Li, Chen, Chai, and Xiong]{li2024gilot}
Xuhong Li, Jiamin Chen, Yekun Chai, and Haoyi Xiong.
\newblock Gi{LOT}: Interpreting generative language models via optimal transport.
\newblock In \emph{Forty-first International Conference on Machine Learning}, 2024.
\newblock URL \url{https://openreview.net/forum?id=qKL25sGjxL}.

\bibitem[Lin et~al.(2025)Lin, Gerchanovsky, Akgul, Bauer, Fredrikson, and Wang]{lin2025llmwhispererinconspicuousattack}
Weiran Lin, Anna Gerchanovsky, Omer Akgul, Lujo Bauer, Matt Fredrikson, and Zifan Wang.
\newblock Llm whisperer: An inconspicuous attack to bias llm responses, 2025.
\newblock URL \url{https://arxiv.org/abs/2406.04755}.

\bibitem[Liu et~al.(2025{\natexlab{a}})Liu, Kandpal, and Raffel]{liu2025attribot}
Fengyuan Liu, Nikhil Kandpal, and Colin Raffel.
\newblock Attribot: A bag of tricks for efficiently approximating leave-one-out context attribution.
\newblock In \emph{The Thirteenth International Conference on Learning Representations}, 2025{\natexlab{a}}.
\newblock URL \url{https://openreview.net/forum?id=9kJperA2a4}.

\bibitem[Liu et~al.(2023{\natexlab{a}})Liu, Lin, Hewitt, Paranjape, Bevilacqua, Petroni, and Liang]{liu2023lostmiddlelanguagemodels}
Nelson~F. Liu, Kevin Lin, John Hewitt, Ashwin Paranjape, Michele Bevilacqua, Fabio Petroni, and Percy Liang.
\newblock Lost in the middle: How language models use long contexts, 2023{\natexlab{a}}.
\newblock URL \url{https://arxiv.org/abs/2307.03172}.

\bibitem[Liu et~al.(2025{\natexlab{b}})Liu, Wang, Qin, Lu, Chen, Yang, and Shu]{liu2025adretrieval}
Tongtong Liu, Zhaohui Wang, Meiyue Qin, Zenghui Lu, Xudong Chen, Yuekui Yang, and Peng Shu.
\newblock Real-time ad retrieval via llm-generative commercial intention for sponsored search advertising.
\newblock \emph{arXiv preprint arXiv:2504.01304}, 2025{\natexlab{b}}.

\bibitem[Liu et~al.(2023{\natexlab{b}})Liu, Iter, Xu, Wang, Xu, and Zhu]{liu2023gevalnlgevaluationusing}
Yang Liu, Dan Iter, Yichong Xu, Shuohang Wang, Ruochen Xu, and Chenguang Zhu.
\newblock G-eval: Nlg evaluation using gpt-4 with better human alignment, 2023{\natexlab{b}}.
\newblock URL \url{https://arxiv.org/abs/2303.16634}.

\bibitem[Lundberg and Lee(2017)]{kernelshap}
Scott~M Lundberg and Su-In Lee.
\newblock A unified approach to interpreting model predictions.
\newblock \emph{Advances in neural information processing systems}, 30, 2017.

\bibitem[McDonald(2025)]{seer}
Tracy McDonald.
\newblock {Google AI Overview Study - SEO \& PPC CTR impact}.
\newblock \emph{{Seer Interactive}}, 2 2025.
\newblock URL \url{https://www.seerinteractive.com/insights/ctr-aio}.
\newblock [Online; accessed 2025-12-17].

\bibitem[Michalak et~al.(2013)Michalak, Aadithya, Szczepanski, Ravindran, and Jennings]{michalak2013efficient}
Tomasz~P Michalak, Karthik~V Aadithya, Piotr~L Szczepanski, Balaraman Ravindran, and Nicholas~R Jennings.
\newblock Efficient computation of the shapley value for game-theoretic network centrality.
\newblock \emph{Journal of Artificial Intelligence Research}, 46:\penalty0 607--650, 2013.

\bibitem[Mitchell et~al.(2022)Mitchell, Cooper, Frank, and Holmes]{mitchell2022sampling}
Rory Mitchell, Joshua Cooper, Eibe Frank, and Geoffrey Holmes.
\newblock Sampling permutations for shapley value estimation.
\newblock \emph{Journal of Machine Learning Research}, 23\penalty0 (43):\penalty0 1--46, 2022.

\bibitem[Mordo et~al.(2024)Mordo, Tennenholtz, and Kurland]{mordo2024sponsored}
Tommy Mordo, Moshe Tennenholtz, and Oren Kurland.
\newblock Sponsored question answering.
\newblock In \emph{Proceedings of the 2024 ACM SIGIR International Conference on Theory of Information Retrieval}, pages 167--173, 2024.

\bibitem[Nematov et~al.(2025)Nematov, Kalai, Kuzmenko, Fugagnoli, Sacharidis, Hose, and Sagi]{nematov2025source}
Ikhtiyor Nematov, Tarik Kalai, Elizaveta Kuzmenko, Gabriele Fugagnoli, Dimitris Sacharidis, Katja Hose, and Tomer Sagi.
\newblock Source attribution in retrieval-augmented generation.
\newblock \emph{arXiv preprint arXiv:2507.04480}, 2025.

\bibitem[Nestaas et~al.(2025)Nestaas, Debenedetti, and Tram{\`e}r]{nestaas2025adversarial}
Fredrik Nestaas, Edoardo Debenedetti, and Florian Tram{\`e}r.
\newblock Adversarial search engine optimization for large language models.
\newblock In \emph{The Thirteenth International Conference on Learning Representations}, 2025.
\newblock URL \url{https://openreview.net/forum?id=hkdqxN3c7t}.

\bibitem[Novet and Elias(2025)]{chegg}
Jordan Novet and Jennifer Elias.
\newblock {Chegg sues Google for hurting traffic as it considers alternatives}.
\newblock CNBC, 2 2025.
\newblock URL \url{https://www.cnbc.com/2025/02/24/chegg-sues-google-for-hurting-traffic-as-it-considers-alternatives.html}.
\newblock [Online; accessed 2025-10-18].

\bibitem[{OpenAI}(2025)]{openai2025gpt41}
{OpenAI}.
\newblock Introducing {GPT}-4.1 in the {API}.
\newblock \url{https://openai.com/index/gpt-4-1/}, 2025.
\newblock Accessed: 2025-10-07.

\bibitem[Park et~al.(2023)Park, Georgiev, Ilyas, Leclerc, and Madry]{trak}
Sung~Min Park, Kristian Georgiev, Andrew Ilyas, Guillaume Leclerc, and Aleksander Madry.
\newblock {TRAK}: Attributing model behavior at scale.
\newblock In Andreas Krause, Emma Brunskill, Kyunghyun Cho, Barbara Engelhardt, Sivan Sabato, and Jonathan Scarlett, editors, \emph{Proceedings of the 40th International Conference on Machine Learning}, volume 202 of \emph{Proceedings of Machine Learning Research}, pages 27074--27113. PMLR, 23--29 Jul 2023.
\newblock URL \url{https://proceedings.mlr.press/v202/park23c.html}.

\bibitem[Perez(2023)]{helena}
Sarah Perez.
\newblock {News publisher files class action antitrust suit against Google, citing AI's harms to their bottom line}.
\newblock \emph{{TechCrunch}}, 12 2023.
\newblock URL \url{https://techcrunch.com/2023/12/15/news-publisher-files-class-action-antitrust-suit-against-google-citing-ais-harms-to-their-bottom-line/}.
\newblock Online; accessed 2025-10-18.

\bibitem[Perplexity~AI(2022)]{perplexity}
Inc. Perplexity~AI.
\newblock Perplexity ai: Answer engine.
\newblock Website / Service, 2022.
\newblock URL \url{https://perplexity.ai}.

\bibitem[Press(2025)]{anthropicbillions2025}
The~Associated Press.
\newblock Anthropic to pay \$1.5 billion to settle authors' copyright lawsuit, 2025.
\newblock URL \url{https://www.cbsnews.com/news/anthropic-copyright-lawsuit-class-action-settlement-authors-1-5-billion/}.

\bibitem[Qi et~al.(2024)Qi, Sarti, Fernández, and Bisazza]{qi2024model}
Jirui Qi, Gabriele Sarti, Raquel Fernández, and Arianna Bisazza.
\newblock Model internals-based answer attribution for trustworthy retrieval-augmented generation.
\newblock In \emph{Proceedings of the 2024 Conference on Empirical Methods in Natural Language Processing}, page 6037–6053. Association for Computational Linguistics, 2024.
\newblock \doi{10.18653/v1/2024.emnlp-main.347}.
\newblock URL \url{http://dx.doi.org/10.18653/v1/2024.emnlp-main.347}.

\bibitem[Ribeiro et~al.(2016)Ribeiro, Singh, and Guestrin]{lime}
Marco~Tulio Ribeiro, Sameer Singh, and Carlos Guestrin.
\newblock ``why should i trust you?'' explaining the predictions of any classifier.
\newblock In \emph{Proceedings of the 22nd ACM SIGKDD international conference on knowledge discovery and data mining}, pages 1135--1144, 2016.

\bibitem[Ritchie(2025)]{thecurrent2025}
Tom Ritchie.
\newblock Ai overviews: How are publishers adapting to the rise of clickless search?, 2025.
\newblock URL \url{https://www.thecurrent.com/marketing-strategy-ai-overviews-publishers-rise-clickless-search}.

\bibitem[Salinas and Morstatter(2024)]{salinas2024butterflyeffectalteringprompts}
Abel Salinas and Fred Morstatter.
\newblock The butterfly effect of altering prompts: How small changes and jailbreaks affect large language model performance, 2024.
\newblock URL \url{https://arxiv.org/abs/2401.03729}.

\bibitem[Santhanam et~al.(2021)Santhanam, Khattab, Saad-Falcon, Potts, and Zaharia]{colbertv2}
Keshav Santhanam, Omar Khattab, Jon Saad-Falcon, Christopher Potts, and Matei Zaharia.
\newblock Colbertv2: Effective and efficient retrieval via lightweight late interaction.
\newblock \emph{arXiv preprint arXiv:2112.01488}, 2021.

\bibitem[Shapley(1953)]{shapley1951notes}
Lloyd~S Shapley.
\newblock A value for n-person games.
\newblock In \emph{Contributions to the theory of games}, volume~2, pages 307--317. Princeton University Press, 1953.
\newblock URL \url{https://www.rand.org/pubs/research_memoranda/RM0670.html}.

\bibitem[Shapley et~al.(1953)]{shapley1953value}
Lloyd~S Shapley et~al.
\newblock \emph{A value for n-person games}.
\newblock Princeton University Press Princeton, 1953.

\bibitem[SimilarWeb(2025)]{similarweb}
SimilarWeb, 2025.
\newblock URL \url{https://www.similarweb.com/}.

\bibitem[Sommerfeld et~al.(2025)Sommerfeld, McCurry, and Harrington]{bain2}
Natasha Sommerfeld, Megan McCurry, and Doug Harrington.
\newblock {Goodbye Clicks, Hello AI: Zero-Click Search Redefines Marketing}.
\newblock \emph{Bain \& Company}, 2 2025.
\newblock URL \url{https://www.bain.com/insights/goodbye-clicks-hello-ai-zero-click-search-redefines-marketing/}.
\newblock Online; accessed 2025-12-04.

\bibitem[Strauss et~al.(2025)Strauss, Yang, O'Reilly, Rosenblat, and Moure]{strauss2025attribution}
Ilan Strauss, Jangho Yang, Tim O'Reilly, Sruly Rosenblat, and Isobel Moure.
\newblock The attribution crisis in llm search results.
\newblock \emph{arXiv preprint arXiv:2508.00838}, 2025.

\bibitem[{The New York Times Company v. Microsoft Corporation et al.}(2023)]{nyt_v_openai_2023}
{The New York Times Company v. Microsoft Corporation et al.}
\newblock No. 1:23-cv-11195, U.S. District Court, Southern District of New York, 2023.
\newblock URL \url{https://dockets.justia.com/docket/new-york/nysdce/1:2023cv11195/612697}.

\bibitem[Trivedi et~al.(2022)Trivedi, Balasubramanian, Khot, and Sabharwal]{trivedi2022musiquemultihopquestionssinglehop}
Harsh Trivedi, Niranjan Balasubramanian, Tushar Khot, and Ashish Sabharwal.
\newblock Musique: Multihop questions via single-hop question composition, 2022.
\newblock URL \url{https://arxiv.org/abs/2108.00573}.

\bibitem[Wang et~al.(2024)Wang, Deng, Chiba-Okabe, Barak, and Su]{wang2024economic}
Jiachen~T Wang, Zhun Deng, Hiroaki Chiba-Okabe, Boaz Barak, and Weijie~J Su.
\newblock An economic solution to copyright challenges of generative ai.
\newblock \emph{arXiv preprint arXiv:2404.13964}, 2024.

\bibitem[Wang et~al.(2025{\natexlab{a}})Wang, Mittal, Song, and Jia]{datashapley2025}
Jiachen~T. Wang, Prateek Mittal, Dawn Song, and Ruoxi Jia.
\newblock Data shapley in one training run.
\newblock In \emph{The Thirteenth International Conference on Learning Representations}, 2025{\natexlab{a}}.

\bibitem[Wang et~al.(2025{\natexlab{b}})Wang, Zou, Geng, and Jia]{tracllm}
Yanting Wang, Wei Zou, Runpeng Geng, and Jinyuan Jia.
\newblock Tracllm: A generic framework for attributing long context llms, 2025{\natexlab{b}}.
\newblock URL \url{https://arxiv.org/abs/2506.04202}.

\bibitem[Wicklin(2023)]{wicklin2023interpret}
Rick Wicklin.
\newblock How to interpret spearman and {Kendall} correlation coefficients.
\newblock The DO Loop Blog, SAS Institute, April 2023.
\newblock URL \url{https://blogs.sas.com/content/iml/2023/04/05/interpret-spearman-kendall-corr.html}.

\bibitem[Xiao et~al.(2025)Xiao, Zhu, Samyoun, Zhang, Wang, and Du]{tokenshapley}
Yingtai Xiao, Yuqing Zhu, Sirat Samyoun, Wanrong Zhang, Jiachen~T. Wang, and Jian Du.
\newblock Tokenshapley: Token level context attribution with shapley value, 2025.
\newblock URL \url{https://arxiv.org/abs/2507.05261}.

\bibitem[Yang et~al.(2018)Yang, Qi, Zhang, Bengio, Cohen, Salakhutdinov, and Manning]{yang2018hotpotqadatasetdiverseexplainable}
Zhilin Yang, Peng Qi, Saizheng Zhang, Yoshua Bengio, William~W. Cohen, Ruslan Salakhutdinov, and Christopher~D. Manning.
\newblock Hotpotqa: A dataset for diverse, explainable multi-hop question answering, 2018.
\newblock URL \url{https://arxiv.org/abs/1809.09600}.

\bibitem[Zeff and Aronson(1999)]{zeff1999advertising}
Robbin~Lee Zeff and Bradley Aronson.
\newblock \emph{Advertising on the Internet}.
\newblock John Wiley \& Sons, Inc., 1999.

\bibitem[Zheng et~al.(2023)Zheng, Chiang, Sheng, Zhuang, Wu, Zhuang, Lin, Li, Li, Xing, Zhang, Gonzalez, and Stoica]{zheng2023judgingllmasajudgemtbenchchatbot}
Lianmin Zheng, Wei-Lin Chiang, Ying Sheng, Siyuan Zhuang, Zhanghao Wu, Yonghao Zhuang, Zi~Lin, Zhuohan Li, Dacheng Li, Eric~P. Xing, Hao Zhang, Joseph~E. Gonzalez, and Ion Stoica.
\newblock Judging llm-as-a-judge with mt-bench and chatbot arena, 2023.
\newblock URL \url{https://arxiv.org/abs/2306.05685}.

\end{thebibliography}

\section*{\LARGE Appendix}
 \appendix
\crefalias{section}{appendix}
\crefname{appendix}{appendix}{appendices}
\Crefname{appendix}{Appendix}{Appendices}

\section{Additional Related Work}
\label{sec:related}

\paragraph{LLM and Online Advertisement.}
LLM has been widely used in online advertisement systems due to its strong natural language understanding and generation capabilities~\citep{feizi2023online,liu2025adretrieval,duetting2024mechanism}.
Many prior works have explored the mechanism design and auction design for LLM-based advertisement systems~\citep{hajiaghayi2024ad, duetting2024mechanism, banchio2024ads, mordo2024sponsored, auction-llm}.
The setting, however, is \textit{orthogonal} to our work, where the focus is on the interaction between the advertisers and the platform that serves the ads, where the advertisers are typically bidding for user attention.
Our setting focuses on the interaction between information providers and generative search providers, where the information providers passively provide information to the service providers and usually only display ads from a third-party advertisement platform.
In fact, the two settings can be considered complementary, where the fair attribution scores from \name can be used as a passive ``bid'' for the information providers to participate in the auction-based advertisement systems.

\paragraph{Attribution Problem in Machine Learning.}
The attribution problem has been extensively studied in the broader machine learning community. For training-time attribution, Data model~\citep{datamodel} and TRAK~\citep{trak} work by learning a prediction model on the impact of each training data point on the model's performance, while Data Shapley approach~\citep{datashapley2019,datashapley2025, wang2024economic} applies the Shapley value concept to quantify the contribution of each training data point to the model's performance.
For inference-time attribution, LIME~\citep{lime} works by learning a local surrogate model to explain the model's prediction under the pertubation the input features,
while Kernel SHAP~\citep{kernelshap} applies the Shapley value concept to quantify the contribution of each input feature to the model's prediction.
Influence function method~\citep{influencefunction} traces the attribution across the inference-training pipeline and
attribute the model's prediction to a specific subset of training data.

\ignore{
	\paragraph{Shapley Value.}
	The Shapley value has established a concrete role in machine learning as a principled attribution method, valued for its fairness properties despite the massive computational inefficiency of exact computation. It has been applied in a variety of settings in ML, like feature selection, explanability, data valuation and more. \sara{edit}
	Exact computation of Shapley values is infeasible in most practical settings, and a range of approximation methods have been developed to address this challenge. Monte Carlo approaches estimate Shapley values by sampling permutations and averaging marginal contributions, but these methods suffer from slow convergence and high variance[]. To address these limitations, a number of refinements have been proposed, including structured sampling, variance-reduction techniques such as antithetic sampling, and utility-bounding strategies[]. Despite these advances, approximation algorithms inevitably sacrifice some of the fairness guarantees and interpretability that motivate the Shapley framework in the first place \sara{edit}.
}

\section{Reward Allocation Mechanisms}
\label{sec:incentive-model}

We envision reward allocation mechanisms (\name or others) could be used in various ways to compensate content providers.

\paragraph{Direct Payment based on Fair Attribution.}
One straightforward application of \name is to use the attributed values as a ratio to allocate a fixed budget to information providers based on their contributions to the final answers.
This budget can be funded by either the users (e.g. through a subscription fee) or the generative search providers (e.g. through a fraction of their own ads revenue).
A primary challenge would be establishing a payment channel between search providers and content providers; this may be feasible in domain-specific scenarios with limited content providers (e.g., academic publishers, news sites).

\paragraph{Advertisement Proxy based on Fair Attribution.}
Another possibility is to use a generative search engine to forward advertisements to viewers.
That is, the generative search engine can detect the displayed advertisement on the search result pages.
Once the attributed values provided by \name are obtained,
search providers can use the attributed values either as a probability distribution or an auction bid to allocate the advertisement slots to information providers,
then show the corresponding advertisements to the users.
Hence, content providers can still earn  advertisement revenue.
This model is (relatively) more backwards compatible with today's web advertisement ecosystem. One potential downside is that advertisements displayed alongside LLM-generated answers may be less effective than in their original form, on their own webpages.

\paragraph{Ad Auction Mechanism based on \name Attribution.}
Finally, \name could be combined with other auction-based mechanisms for advertisement allocation.
\citet{hajiaghayi2024ad} proposed an auction-based mechanism for LLM-generated answers, where each advertiser bids on the opportunity to influence the final response.
In their paper, a key technique is to compute the ``adjusted bids'' for each advertiser based on their bid and also an ``attribution score'' that is assumed to be available and linearly related to the click-through rate (CTR).
The core of their mechanism is a probabilistic second-price auction based on the adjusted bids. (Crucially, this setting requires attribution to be computed by a third party (the advertiser or platform) without access to model internals, making black-box attribution a structural necessity rather than merely a practical convenience.)
\name could be used to compute the attribution score for each advertiser based on their contribution to the LLM's answer.

\section{Coverage-Game Derivation}
\label{app:coverage-games}

This section develops coverage games as a mathematical device for deriving the
max-game Shapley formula used in \Cref{sec:shapley-computation}.  The geometric
language is only an intuition: one may picture players as covering parts of an
abstract space, such as regions of a map or pieces of answer-relevant evidence
in a generative search answer.

\subsection{General Coverage Game}

Let $(\Omega,\mathcal{F},\mu)$ be a measurable space.  Each player
$i\in[m]$ is associated with a measurable set $A_i\subseteq\Omega$ (i.e., their individual coverage set).
Let $\rho(x)\ge 0$ be an integrable value density.  For an index set $I\subseteq [m]$, define the local coverage count
\[
h_I(x)=\sum_{i\in I}\mathbf{1}\{x\in A_i\}.
\]
Given a coverage curve $g:\mathbb{Z}_{\ge 0}\to\mathbb{R}_{\ge 0}$ with $g(0)=0$, define the coverage-game utility
\begin{align}
U_g(I)
=
\int_{\Omega}\rho(x)\,g(h_I(x))\,d\mu(x).
\label{eq:coverage-game-utility}
\end{align}

The intuition of the utility is as follows: the players in subset $I$ cover the union of their sets. For each covered ``local point'' $x \in \bigcup_{i\in I} A_i$, it has a value of $\rho(x)$. Moreover, if this point $x$ is covered by a certain number of players, denoted by $h_I(x)$, the multiplicative factor $g(h_I(x))$ adjusts the reward accordingly. So each point contributes $\rho(x)\cdot g(h_I(x))$ to the utility.

Moreover, the condition $g(0)=0$ makes the empty local coalition contribute zero.  It is
often convenient, though not required for the derivation below, to choose $g$
nondecreasing.  Concavity of $g$ is another optional modeling choice, encoding
diminishing returns as more players cover the same point.

The coverage curve is the only local rule.  Max coverage,
$g(t)=\mathbf{1}\{t>0\}$, assigns value once at least one selected player
covers the point.  Additive coverage, $g(t)=t$, assigns one unit of local value
per selected covering player.  Other choices, such as $g(t)=\min(t,k)$ or
$g(t)=\sum_{r=1}^t\alpha_r$ for non-increasing $\alpha_r\ge 0$, interpolate
between these extremes by changing how much value is assigned to multiply
covered points.

\begin{figure}[t]
    \centering
    \resizebox{\textwidth}{!}{\definecolor{icgblue}{RGB}{44,99,177}
\definecolor{icggreen}{RGB}{48,132,93}
\definecolor{icgorange}{RGB}{210,116,45}
\definecolor{icggray}{RGB}{68,68,68}

\begin{tikzpicture}[x=1cm,y=1cm,every node/.style={font=\footnotesize}]
    \path[use as bounding box] (-0.10,-0.98) rectangle (12.10,5.48);

    \fill[white,rounded corners=2pt]
        (0,0) rectangle (12.0,5.25);

    \begin{scope}[rotate around={-16:(4.28,2.72)}]
        \filldraw[fill=icgblue!62,fill opacity=0.30,draw=icgblue!80!black,
                  draw opacity=0.98,line width=1.05pt]
            (4.28,2.72) ellipse[x radius=3.08,y radius=1.28];
    \end{scope}

    \filldraw[fill=icggreen!60,fill opacity=0.30,draw=icggreen!70!black,
              draw opacity=0.98,line width=1.05pt]
        (3.28,1.06)
        .. controls (4.52,0.38) and (7.70,0.45) .. (8.95,1.48)
        .. controls (10.05,2.38) and (9.36,4.08) .. (7.55,4.12)
        .. controls (5.92,4.16) and (4.62,3.48) .. (3.76,2.46)
        .. controls (3.30,1.92) and (2.88,1.39) .. cycle;

    \begin{scope}[rotate around={15:(7.15,3.02)}]
        \filldraw[fill=icgorange!70,fill opacity=0.31,draw=icgorange!82!black,
                  draw opacity=0.98,line width=1.05pt,rounded corners=8pt]
            (5.68,2.03) rectangle (8.62,4.04);
    \end{scope}

    \node[icgblue!78!black,font=\bfseries] at (2.15,3.38) {$A_1$};
    \node[icggreen!56!black,font=\bfseries] at (8.86,2.84) {$A_2$};
    \node[icgorange!76!black,font=\bfseries] at (7.86,3.94) {$A_3$};

    \fill[black] (6.35,2.75) circle (1.65pt);
    \node[fill=white,fill opacity=0.90,text opacity=1,inner sep=1.7pt,
          rounded corners=1pt,anchor=west]
        at (6.52,2.75) {$h(x)=3$};

    \fill[black] (4.78,1.86) circle (1.65pt);
    \node[fill=white,fill opacity=0.90,text opacity=1,inner sep=1.7pt,
          rounded corners=1pt,anchor=east]
        at (4.60,1.86) {$h(x)=2$};

    \fill[black] (8.58,1.78) circle (1.65pt);
    \node[fill=white,fill opacity=0.90,text opacity=1,inner sep=1.7pt,
          rounded corners=1pt,anchor=west]
        at (8.75,1.78) {$h(x)=1$};

    \draw[draw=icggray!45,rounded corners=2pt,line width=0.7pt]
        (0,0) rectangle (12.0,5.25);
    \node[anchor=north west,icggray,font=\footnotesize\bfseries]
        at (0.22,5.03) {$\Omega$};

    \draw[draw=icggray!28,fill=icggray!4,rounded corners=2pt,line width=0.55pt]
        (0,-0.86) rectangle (12.0,-0.18);
    \draw[icggray!25] (4.35,-0.78) -- (4.35,-0.26);
    \draw[icggray!25] (7.80,-0.78) -- (7.80,-0.26);
    \node[anchor=center,font=\scriptsize] at (2.18,-0.52)
        {$h(x)=|\{i:x\in A_i\}|$};
    \node[anchor=center,font=\scriptsize] at (6.08,-0.52)
        {value $g(h(x))$};
    \node[anchor=center,font=\scriptsize] at (9.90,-0.52)
        {share $g(h(x))/h(x)$};
\end{tikzpicture}}
    \caption{An illustrative two-dimensional coverage game with three players, drawn under uniform value density for simplicity. Each player $s_i$ is represented by a measurable set $A_i\subseteq\Omega$. At a point $x$, the local count $h(x)$ is the number of selected sets containing that point; the coverage curve $g$ determines the local value $g(h(x))$.}
    \label{fig:app-information-coverage-game}
\end{figure}

\begin{theorem}[Pointwise Shapley value for coverage games]
\label{thm:coverage-game-shapley}
Let $h(x)=h_{[m]}(x)$ be the total number of player sets containing point $x$.
For the utility in \Cref{eq:coverage-game-utility}, the Shapley value of
player $i$ is
\begin{align}
\phi_i^{U_g}
=
\int_{A_i}
\rho(x)\frac{g(h(x))}{h(x)}\,d\mu(x),
\label{eq:coverage-game-shapley}
\end{align}
where the integrand is taken to be zero when $h(x)=0$.
\end{theorem}

\begin{proof}
Fix a point $x\in\Omega$ and consider the local game
$u_x(I)=\rho(x)g(h_I(x))$. If $x\notin A_i$, then player $i$ is a null player
for $u_x$ and its local Shapley value is zero. If $x\in A_i$, let
$C(x)=\{j:x\in A_j\}$ and $h(x)=|C(x)|$. In a uniformly random permutation of
players, the rank of $i$ among the $h(x)$ players in $C(x)$ is uniform on
$\{1,\ldots,h(x)\}$. If this rank is $r$, then exactly $r-1$ covering players
appeared earlier, and the local marginal contribution of player $i$ is
\[
\rho(x)\left(g(r)-g(r-1)\right).
\]
Therefore the expected local marginal contribution is
\[
\frac{\rho(x)}{h(x)}
\sum_{r=1}^{h(x)}\left(g(r)-g(r-1)\right)
=
\rho(x)\frac{g(h(x))}{h(x)},
\]
using $g(0)=0$. Integrating these pointwise expected marginal contributions over $\Omega$ gives \Cref{eq:coverage-game-shapley}.
\end{proof}

The theorem is the main reason to introduce coverage games: once the utility is
written in the form of \Cref{eq:coverage-game-utility}, the Shapley value has an
immediate pointwise expression.  The total local value at $x$ is
$\rho(x)g(h(x))$, and Shapley symmetry divides that value equally among the
$h(x)$ players whose sets contain $x$.

\subsection{The Max Game as a One-Dimensional Coverage Game}
\label{app:max-game-coverage}

The key-point maximization game used by \name is the one-dimensional
max-coverage special case.  Let $\Omega=\mathbb{R}_{\ge 0}$ with Lebesgue
measure and uniform density $\rho(x)=1$.  A player with non-negative score
$v_i$ is represented by the interval
\[
A_i=[0,v_i].
\]
Use the max-coverage curve $g(t)=\mathbf{1}\{t>0\}$. For any coalition $I\subseteq [m]$, \Cref{eq:coverage-game-utility} becomes
\begin{align}
U_g(I)
&=
\int_0^\infty
\mathbf{1}\{\exists i\in I:x\le v_i\}\,dx
=
\max_{i\in I}v_i,
\label{eq:max-game-as-coverage}
\end{align}
with the convention that the maximum over the empty set is zero. Thus, the
usual max utility is exactly the length of the union of one-dimensional
intervals. This identity is algebraic; the interval view is useful because it
gives a direct sorted-scan Shapley formula.

\begin{figure}[t]
    \centering
    \resizebox{0.95\textwidth}{!}{\definecolor{covblue}{RGB}{44,99,177}
\definecolor{covgreen}{RGB}{48,132,93}
\definecolor{covorange}{RGB}{210,116,45}
\definecolor{covgray}{RGB}{68,68,68}
\definecolor{covviolet}{RGB}{112,82,170}

\begin{tikzpicture}[x=1cm,y=1cm,every node/.style={font=\footnotesize}]
    \path[use as bounding box] (-0.35,-1.25) rectangle (9.85,3.35);

    \node[anchor=east,covgray] at (-0.12,2.55) {$s_1$};
    \filldraw[fill=covblue!28,draw=covblue!85!black,line width=0.85pt,rounded corners=1pt]
        (0,2.37) rectangle (8.00,2.73);
    \node[covblue!80!black] at (3.95,2.55) {$A_1=[0,v_1]$};
    \node[anchor=west,covblue!80!black] at (8.10,2.55) {$v_1=1.0$};

    \node[anchor=east,covgray] at (-0.12,1.80) {$s_2$};
    \filldraw[fill=covgreen!30,draw=covgreen!75!black,line width=0.85pt,rounded corners=1pt]
        (0,1.62) rectangle (5.60,1.98);
    \node[covgreen!60!black] at (2.85,1.80) {$A_2=[0,v_2]$};
    \node[anchor=west,covgreen!60!black] at (5.70,1.80) {$v_2=0.7$};

    \node[anchor=east,covgray] at (-0.12,1.05) {$s_3$};
    \filldraw[fill=covorange!32,draw=covorange!85!black,line width=0.85pt,rounded corners=1pt]
        (0,0.87) rectangle (3.20,1.23);
    \node[covorange!75!black] at (1.60,1.05) {$A_3=[0,v_3]$};
    \node[anchor=west,covorange!75!black] at (3.30,1.05) {$v_3=0.4$};

    \draw[<->,covblue!80!black,line width=0.45pt]
        (0,3.03) -- (8.00,3.03);
    \node[fill=white,inner xsep=2pt,text=covblue!80!black] at (4.00,3.03)
        {relevance score $v_1=1.0$};

    \draw[->,covgray,line width=0.65pt] (0,0) -- (9.25,0) node[right] {support axis $x$};
    \node[anchor=north,covgray] at (0,-0.03) {$0$};

    \foreach \x/\lab in {3.20/$v_3$,5.60/$v_2$,8.00/$v_1$} {
        \draw[densely dashed,covgray!55] (\x,-0.95) -- (\x,2.95);
        \node[anchor=north,covgray] at (\x,-0.03) {\lab};
    }

    \filldraw[fill=covviolet!20,draw=covgray!30,line width=0.35pt]
        (0,-0.95) rectangle (3.20,-0.53);
    \filldraw[fill=covblue!18,draw=covgray!30,line width=0.35pt]
        (3.20,-0.95) rectangle (5.60,-0.53);
    \filldraw[fill=covgreen!16,draw=covgray!30,line width=0.35pt]
        (5.60,-0.95) rectangle (8.00,-0.53);
    \filldraw[fill=covgray!10,draw=covgray!30,line width=0.35pt]
        (8.00,-0.95) rectangle (9.20,-0.53);

    \node at (1.60,-0.74) {$h(x)=3$};
    \node at (4.40,-0.74) {$h(x)=2$};
    \node at (6.80,-0.74) {$h(x)=1$};
    \node at (8.60,-0.74) {$h(x)=0$};

    \node[anchor=west,covgray,font=\scriptsize] at (0,-1.16)
        {$h(x)$ is the number of source intervals covering point $x$};
\end{tikzpicture}}
    \caption{The one-dimensional max-coverage representation for one max game with three players. Player $s_i$ is represented by the interval $A_i=[0,v_i]$. The interval endpoints partition the axis into regions with coverage counts $h(x)=3,2,1,0$; under max coverage, Shapley attribution divides each covered region equally among the players covering it.}
    \label{fig:app-coverage-game-1d}
\end{figure}

\begin{corollary}[Exact Shapley computation for the maximization game]
\label{cor:max-game-shapley}
Consider the maximization game $U_{\mathsf{Max}}(I)=\max_{i\in I}v_i$ with non-negative scores $v_1,\ldots,v_m$ and $U_{\mathsf{Max}}(\emptyset)=0$. Sort the scores as $0\le v_1\le\cdots\le v_m$ and set $v_0=0$. The Shapley value of the player with sorted rank $r$ is
\begin{align}
\phi_r^{\mathsf{Max}}
=
\sum_{\ell=1}^{r}\frac{v_\ell-v_{\ell-1}}{m-\ell+1}.
\label{eq:max-game-sorted-shapley}
\end{align}
Therefore all exact Shapley values for the maximization game can be computed by sorting once and scanning the sorted endpoints in $O(m\log m)$ time.
\end{corollary}

\begin{proof}
By \Cref{thm:coverage-game-shapley}, player $i$ receives
\begin{align}
\phi_i^{\mathsf{Max}}
=
\int_0^{v_i}\frac{1}{h(x)}\,dx,
\qquad
h(x)=|\{j:x\le v_j\}|.
\label{eq:max-game-integral-shapley}
\end{align}
The interval segment $(v_{\ell-1},v_\ell]$ is covered by exactly $m-\ell+1$
players, namely the players with sorted rank at least $\ell$. The player with
sorted rank $r$ covers precisely the segments with $\ell\le r$, which gives
\Cref{eq:max-game-sorted-shapley}. The algorithmic claim follows because the
formula requires one sort and one linear scan. Ties require no special case: if
$v_\ell=v_{\ell-1}$, the corresponding segment has zero length and contributes
zero.
\end{proof}

\paragraph{Proof of Proposition~\ref{prop:maxshapley-complexity}.}
Algorithm~\ref{alg:full-algorithm} has two stages.
In Stage I, the non-repeated prompts for keypoint decomposition, distillation, and related bookkeeping contribute $O(T|\Psi|)$ FLOPs under the stated token-cost model; we also include in $T$ the generated outputs of the source-keypoint scoring calls.
For relevance scoring, the algorithm makes one source-keypoint scoring call for every pair $(s_i,p_j)$, hence $nm$ calls.
Each such call has input length at most $L$, including the fixed prompt, query/answer/keypoint context, and source text, so these calls contribute $O(nmL|\Psi|)$ FLOPs.
Thus the LLM-related cost is $O((T+nmL)|\Psi|)$.

In Stage II, after all scores $v_{i,j}$ are fixed, no further LLM calls are needed.
By Corollary~\ref{cor:max-game-shapley}, each key-point maximization game can be solved exactly by sorting the $m$ source scores and performing one linear scan, costing $O(m\log m)$ arithmetic operations.
Aggregating the weighted per-keypoint Shapley values across sources costs an additional $O(m)$ per key point, which is dominated by sorting.
Across $n$ key points, the non-LLM arithmetic cost is therefore $O(nm\log m)$.
Combining the two stages gives $O((T+nmL)|\Psi|+nm\log m)$.

\subsection{Rewarding Corroborating Information with the Coverage-Game View}

The max game used by \name is deliberately conservative: once a key point is
well supported by one source, an additional source supporting the same point
does not increase the key-point utility.  This is a natural default when the
goal is to identify which sources are sufficient for the answer, and it also
avoids over-rewarding near-duplicate documents.

However, some generative search credit-attribution settings may want to value corroboration.  If two
independent sources support the same key point, the second source can still be
useful: it may increase confidence, reduce the risk that the answer rests on a
single fragile source, or provide independent confirmation for a contested
claim.  The coverage-game view gives a simple way to express this choice.  We
can replace the max-coverage curve $g(t)=\mathbf{1}\{t>0\}$ with a curve that
continues to grow after the first covering source, for example
\[
g_\lambda(t)=\mathbf{1}\{t>0\}\bigl(1+\lambda(t-1)\bigr),
\qquad \lambda\in[0,1].
\]
Here $\lambda=0$ recovers the max game, while $\lambda=1$ gives additive
coverage.  Intermediate values reward corroboration without treating every
additional source as equally valuable as the first one.  A concave or capped
curve, such as $g(t)=\min\{1+\lambda(t-1),c\}$, can further limit the reward
from many redundant sources.

Importantly, this modification does not require a new Shapley derivation.  Once
the utility is written as a coverage game, \Cref{thm:coverage-game-shapley}
still applies: the local value at a point is $\rho(x)g(h(x))$, and Shapley
splits that local value equally among the sources covering that point.  Thus,
the choice of $g$ becomes a modeling knob: max coverage rewards sufficiency,
while increasing coverage curves additionally reward corroboration.

For generative search systems, this knob should be used with care.  Corroboration is valuable
when sources are genuinely independent, but it can be misleading when many
retrieved passages are copies, syndications, or minor rewrites of the same
underlying evidence.  In such settings, deduplication, source clustering, or a
concave/capped $g$ may be necessary before using corroboration rewards.  We use
the max game in the main method as the cleanest conservative choice, but the
coverage-game derivation shows how the same framework can support attribution
rules that deliberately reward independent corroborating evidence.

\section{Experimental Setup}

\subsection{Full Evaluation Pipeline}
\label{app:evaluation-pipeline}

Each dataset example consists of a query $q$ and an ordered set of retrieved
sources $S=(s_1,\ldots,s_m)$, where each source is a text snippet. Some datasets
also provide a reference answer $\tilde a$ for evaluation; in our experiments,
this reference answer is available for HotpotQA and MuSiQUE.

For each example, we first run the search LLM $\Psi$ once on the full context:
\[
    a=\Psi(q,S).
\]
The generated answer $a$ is the object being attributed. That is, all source
credit scores are intended to explain the displayed answer $a$, not the reference
answer $\tilde a$.

We then run each attribution method on the same downstream attribution instance:
the query $q$, the sources $S=(s_1,\ldots,s_m)$, and the generated answer $a$.
Different methods obtain scores in different ways. Some baselines repeatedly
evaluate ablated source subsets, while \name uses an attribution LLM $\Psi_A$ as
a scoring tool inside its attribution algorithm. To isolate differences between
attribution methods rather than differences between language models, we use the
same underlying LLM for $\Psi_A$ as for the search LLM $\Psi$ in each model
setting.

We treat FullShapley as the attribution-quality reference. When a dataset
provides a reference answer $\tilde a$, we additionally include $\tilde a$ in the
LLM-judge prompt used to evaluate utilities inside the FullShapley pipeline. This
oracle information is used only to obtain a high-fidelity reference attribution;
it does not change the generated answer $a$ being attributed.

After each method produces source scores, we evaluate them in two ways. First, we
compare the selected or top-ranked sources against human source annotations using
Jaccard index. Second, we compare each method's full ranking of source credits
against FullShapley using Kendall's $\tau_b$.

We repeat this end-to-end evaluation pipeline three times and report aggregate
metrics across runs, reducing variance from stochastic generation and LLM-based
scoring.

\subsection{LLM Prompts}
\label{app:prompts}


In LLM-as-a-judge, we use the Attribution LLM (Figure \ref{fig:setup}) to compute the $\textsf{Judge}$ function, using both GPT-4.1 nano and Haiku 4.5. The LLM $\Psi_A$ receives a query, an LLM-generated response, a subset of the information sources, and, if the ground truth is available, the ground truth to the answer, then produces a score indicating how well that subset answers the query. We designed our prompts to elicit reliable quality assessments while preventing knowledge hallucination---ensuring the model relies solely on provided sources rather than its parametric knowledge. Scores range from 0.0 to 1.0, where 1.0 indicates a perfect answer addressing all parts of the question, and 0.0 indicates an incorrect or unsupported answer. The full prompt is provided in Figure \ref{fig:prompt1}; we provide ground truth only to FullShapley and approximation baselines.

For \name, in addition to the basic LLM-generated response prompt for answering the query with a set of information sources, we used an LLM to break the response down into key points, then to "distill" these key points (filtering out repetitive or redundant key points, see Appendix \ref{app:keypoint_decomp} for details), and then asked the LLM to rank each information source in terms of relevance to each key point, which serves as our value function for \name. We include the full version of the keypoint breakdown prompt in Figure \ref{fig:prompt2}, the distillation prompt in Figure \ref{fig:prompt4}, and the relevancy scoring in Figure \ref{fig:prompt3}.

\begin{figure}[h]
	\centering
\begin{lstlisting}[
		backgroundcolor=\color{black!5},
		frame=single,
		rulecolor=\color{black!30},
		basicstyle=\scriptsize\ttfamily,
		breaklines=true,
		linewidth=\columnwidth
	]
You are an evaluation system that evaluates AI responses to a question. The goal is to determine if the response is correct and well-founded.


EVALUATION FOCUS:
Judge the AI's final answer and how it got to its answer. Correct answers built upon evidence are better than correct guesses. Reasonable inference and pulling together well-founded pieces of information is good. It's crucial to consider what the question is asking as well. 


Question: {input query}
Ground Truth: {input ground_truth, when available}
AI Response: {input LLM-generated response}


Does the AI response fully address the question and convey the same meaning as \"{input LLM-generated response}\" with confirmed information? 


Justification: [Does the AI response fully address the question and convey the same meaning as the ground truth with confirmed information? ]
Score: [Number from 0.0 to 1.0]
\end{lstlisting}
	\caption{Simplified LLM-as-a-judge prompt, FullShapley.}
	\label{fig:prompt1}
\end{figure}

\begin{figure}[h]
	\centering
\begin{lstlisting}[
		backgroundcolor=\color{black!5},
		frame=single,
		rulecolor=\color{black!30},
		basicstyle=\scriptsize\ttfamily,
		breaklines=true,
		linewidth=\columnwidth
	]
You are a document analysis system designed to extract the facts that inform a response to a question.


YOUR PURPOSE:
You should identify the information behind the reasoning of the response. Use how the response answers the question to create the keypoints. The response will be built upon pieces of information pulled together. Your job is to turn each piece of information into a keypoint. 


Question: {query}
Response: {LLM-generated response}


Provide your response in this exact format:
JUSTIFICATION: [Briefly explain the reasoning chain that leads to your keypoints]
KEY POINTS:
[One key point per line]
\end{lstlisting}
	\caption{Sketch, \name keypoint breakdown prompt.}
	\label{fig:prompt2}
\end{figure}

\begin{figure}[t]
	\centering
\begin{lstlisting}[
		backgroundcolor=\color{black!5},
		frame=single,
		rulecolor=\color{black!30},
		basicstyle=\scriptsize\ttfamily,
		breaklines=true,
		linewidth=\columnwidth
	]
You are evaluating if a source document provides substantive support for a specific statement. 

SCORING SCALE (0.0 to 1.0):
- 0.0 = No Support: Source lacks information to support the statement, even if same topic
- 0.3 = Minimal Support: Source has some relevant info but missing key details  
- 0.7 = Substantial Support: Source contains most information needed, minor gaps only
- 1.0 = Complete Support: Source explicitly contains all information to support the statement

RESPONSE FORMAT:
EXPLANATION: [2 sentences on what specific information the source does/doesn't provide]
SCORE: [0.0 to 1.0]

Evaluate how well this source supports the statement:
SOURCE: {source}
STATEMENT: {keypoint}

\end{lstlisting}
	\caption{Keypoint relevance scoring prompt, \name.}
	\label{fig:prompt3}
\end{figure}

\begin{figure}[htbp]
	\centering
\begin{lstlisting}[
		backgroundcolor=\color{black!5},
		frame=single,
		rulecolor=\color{black!30},
		basicstyle=\scriptsize\ttfamily,
		breaklines=true,
		linewidth=\columnwidth
	]
You are evaluating whether a source document provides substantive informational support for a specific statement.
CRITICAL: Being on the same topic is not sufficient. The source must contain specific information that directly supports the statement's claims.
Semantic equivalence or clear logical entailment is allowed. Reasonable and clear interpretation is also allowed - for example, if the statement refers to rectangles and the source refers to squares, that counts as support since the claim logically applies.

SCORING SCALE (0.0 to 1.0):
0.0 = No Support: Source lacks information to support the statement, even if on the same topic.
0.3 = Minimal Support: Source has some relevant information but is missing key details.
0.7 = Substantial Support: Source contains most of the information needed, with only minor gaps.
1.0 = Complete Support: Source explicitly contains all information required to support the statement.

KEY RULE:
Only score based on substantive informational support, not topical similarity.
Statements about what is not mentioned should score 0.0.
\end{lstlisting}
	\caption{Keypoint distillation prompt, \name.}
	\label{fig:prompt4}
\end{figure}

\begin{figure*}[htp]
    \centering
    \begin{minipage}[t]{0.32\textwidth}
        \centering
        \includegraphics[width=\textwidth]{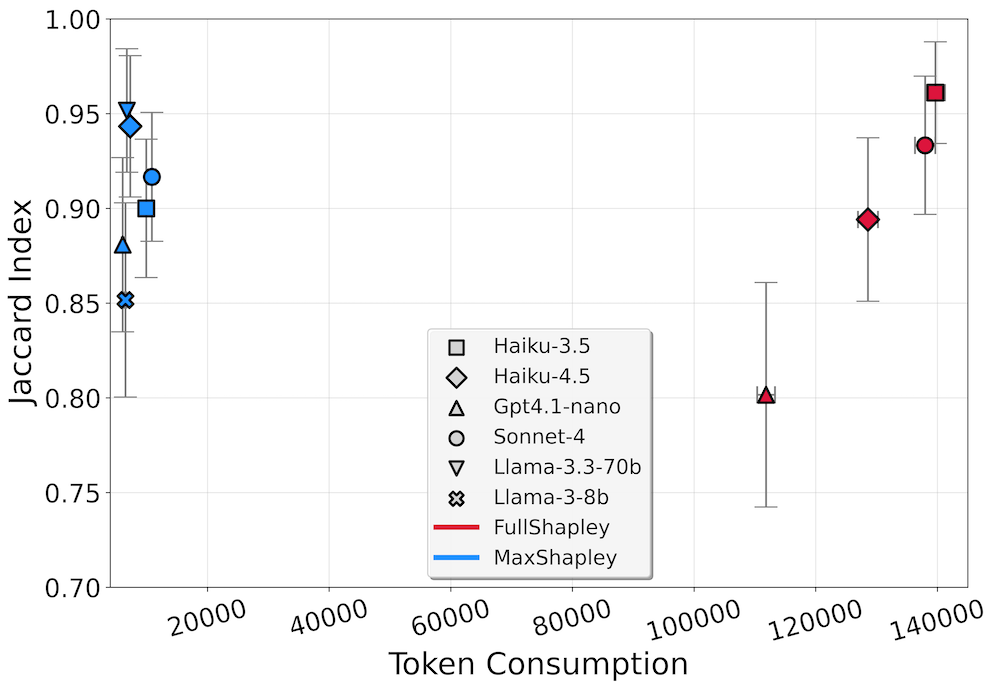}
    \end{minipage}%
    \hfill
    \begin{minipage}[t]{0.32\textwidth}
        \centering
        \includegraphics[width=\textwidth]{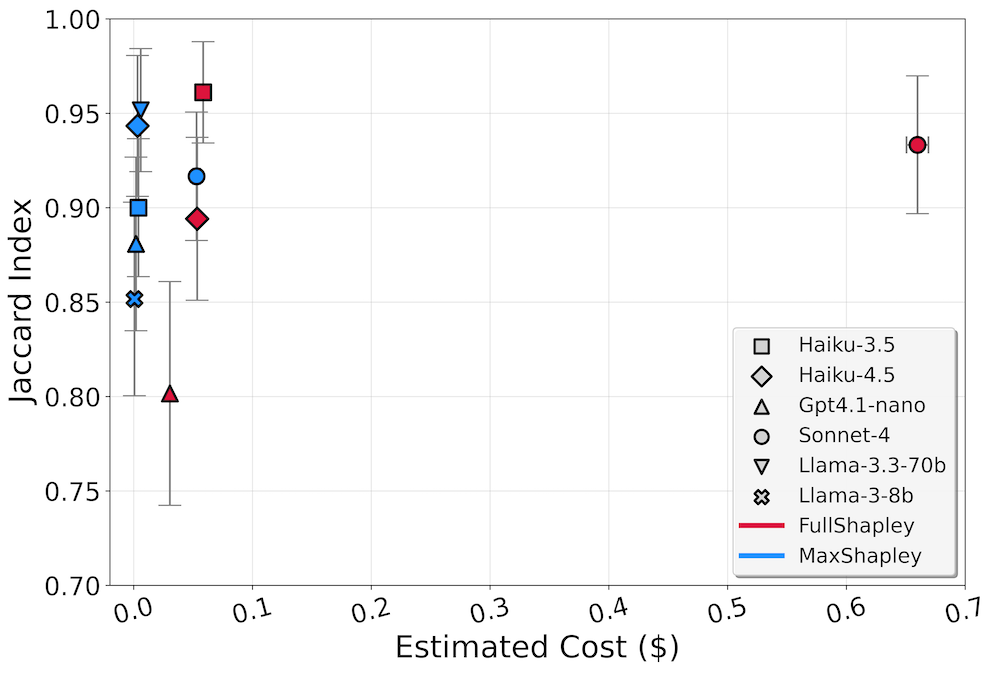}
    \end{minipage}%
    \hfill
    \begin{minipage}[t]{0.32\textwidth}
        \centering
        \includegraphics[width=\textwidth]{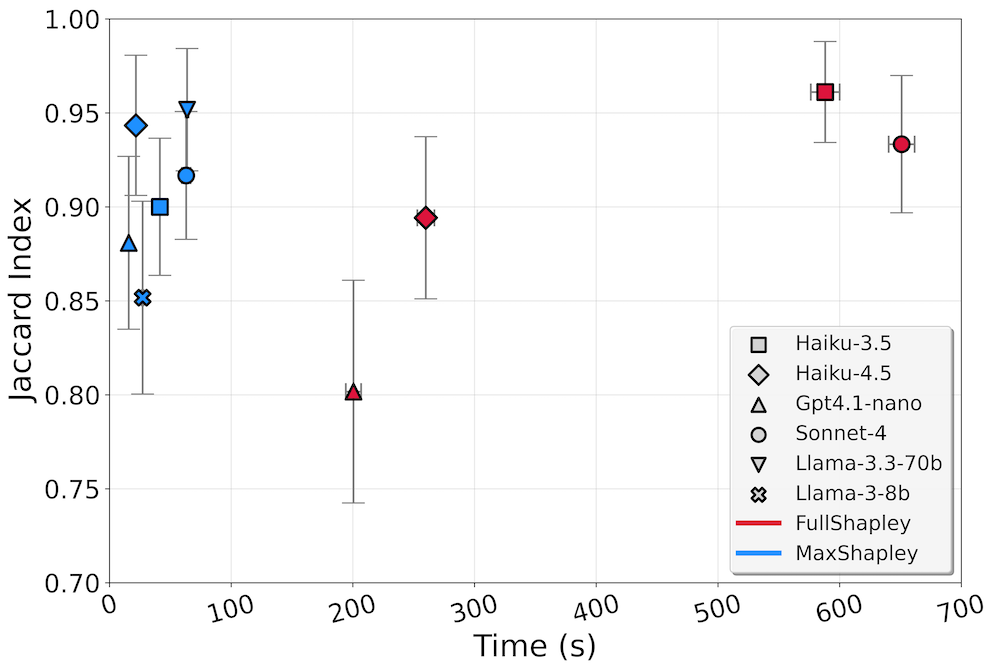}
    \end{minipage}
    \caption{Attribution quality versus (left) token consumption, (center) cost of computation, and (right) computation time across six models and two Shapley algorithms. Haiku 3.5 and 4.5 modestly outperform GPT-4.1 nano and Llama 3 in quality at slightly higher token costs, computation costs, and computation time. Sonnet 4's increased capabilities, costs, and computation time do not translate to quality improvements. Llama 3.3's increased capabilies do result in quality improvements, at cost comparable with Sonnet 4. }
    \label{fig:model_comparison}
\end{figure*}

\subsection{Keypoint Examples}
\label{app:keypoint_examples}

We provide an example of two keypoints resulting from the answering of one query below.

\begin{figure}[htbp]
	\centering
\begin{lstlisting}[
		backgroundcolor=\color{black!5},
		frame=single,
		rulecolor=\color{black!30},
		basicstyle=\scriptsize\ttfamily,
		breaklines=true,
		linewidth=\columnwidth
	]
Query: What government position was held by the woman who portrayed Corliss Archer
in Kiss and Tell?

> Keypoint 1: Shirley Temple portrayed Corliss Archer in the 1945 film.
> Keypoint 2: Shirley Temple served as Chief of Protocol of the United States
\end{lstlisting}
	\caption{Keypoint example.}
\end{figure}

\subsection{Baseline Pseudocode}
\label{app:baselines}

Here, we provide the pseudocode for the baselines that we used in the experiments.
In Algorithm \ref{alg:baselineShapley}, we include the brute-force algorithm for computing Shapley value. While there exist more efficient approximations to the Shapley value, the exact computation is known to have exponential complexity.

\begin{algorithm}[ht]
	\caption{Full Shapley}
	\label{alg:baselineShapley}
	\begin{algorithmic}
	\REQUIRE A value function $V(\cdot)$ and a set of $m$ elements (e.g., information sources) $S = \{s_1, s_2, \dots, s_m\}$.
	\ENSURE Shapley values $\phi_i$ for each $i \in \{1, \dots, m\}$.
	\STATE Initialize $\phi_i \gets 0$ for all $i \in \{1, \dots, m\}$.
	\FOR{$i \in \{1, \dots, m\}$}
		\FOR{$j \in \{0, \dots, m-1\}$}
			\STATE Let $\mathcal{T}_j$ be all subsets of size $j$ from $\{1, \dots, m\} \setminus \{i\}$.
			\FOR{each $T \in \mathcal{T}_j$}
				\STATE $T' \gets T \cup \{i\}$ \COMMENT{Add element $i$ into subset $T$}
				\STATE $v_{\text{with}} \gets V(T')$
				\STATE $v_{\text{without}} \gets V(T)$
				\STATE $\Delta \gets v_{\text{with}} - v_{\text{without}}$ \COMMENT{Marginal contribution of source $i$}
				\STATE $\phi_i \gets \phi_i + \dfrac{\Delta}{\binom{m-1}{j} \cdot m}$
			\ENDFOR
		\ENDFOR
	\ENDFOR
	\RETURN $\{\phi_i\}_{i \in [m]}$
	\end{algorithmic}
\end{algorithm}

Next, we provide pseudocode for the Monte Carlo Approximation of Shapley Value via Sampling, in Algorithm \ref{alg:shapleyMonteCarlo}.

\begin{algorithm}[ht]
	\caption{Monte-Carlo Approximation of Shapley Values via Sampling}
	\label{alg:shapleyMonteCarlo}
	\begin{algorithmic}
	\REQUIRE A value function $V(\cdot)$, number of information sources $m$, and sample size $n$.
	\ENSURE Approximated Shapley values $\phi_i$ for each $i \in \{1, \dots, m\}$.
	\STATE Initialize $\phi_i \gets 0$ for all $i \in \{1, \dots, m\}$.
	\STATE Let $v_\emptyset \gets V(\emptyset)$ \COMMENT{Value of the empty subset}
	\FOR{$r = 1$ to $n$}
		\STATE Sample a random permutation $\pi$ of $\{1, \dots, m\}$ from the uniform distribution.
		\STATE Initialize $T \gets \emptyset$,  $v_{\text{prev}} \gets v_\emptyset$
		\FOR{$i$ in $\pi$}
			\STATE Let $T' \gets T \cup \{i\}$
			\STATE $v_{\text{new}} \gets V(T')$
			\STATE $\Delta \gets v_{\text{new}} - v_{\text{prev}}$ \COMMENT{Marginal contributions}
			\STATE Update $T \gets T'$, $v_{\text{prev}} \gets v_{\text{new}}$
		\ENDFOR
	\ENDFOR
	\RETURN $\{\phi_i\}_{i \in [m]}$
	\end{algorithmic}
\end{algorithm}

\subsection{Dataset Annotation}
\label{app:annotation}

We independently annotated subsets of 100 queries for HotPotQA and MuSiQUE, and 95 queries for MS-MARCO, the maximum number of samples with existing relevance annotations. Each query contains six candidate information sources, selected to include both relevant and irrelevant sources according to the original, noisy dataset labels. Two annotators labeled per-source relevance and then discussed disagreements to reach consensus. Inter-rater reliability before discussion was 85\% for HotPotQA, 86\% for MS-MARCO, and 92\% for MuSiQUE. 

\begin{figure}[htbp]
	\centering
	\begin{minipage}[t]{0.28\linewidth}
	 \centering
	 \includegraphics[width=\textwidth]{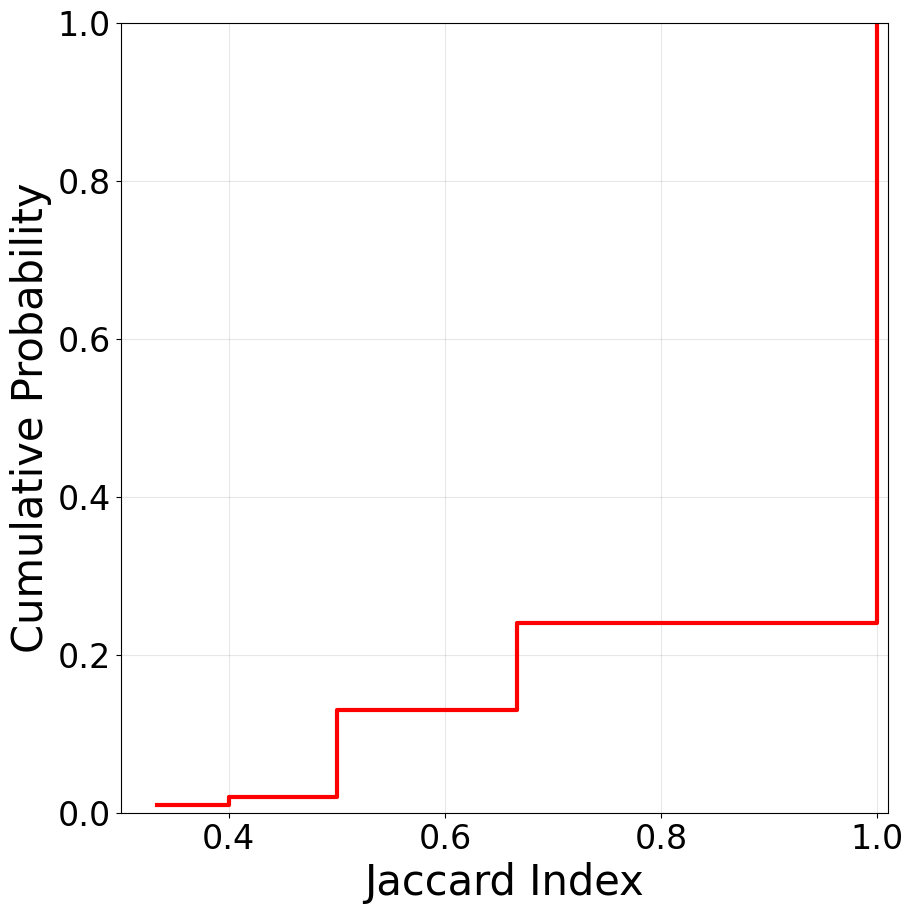}
	\end{minipage}\hspace{0.04\linewidth}%
	\begin{minipage}[t]{0.28\linewidth}
	 \centering
	 \includegraphics[width=\textwidth]{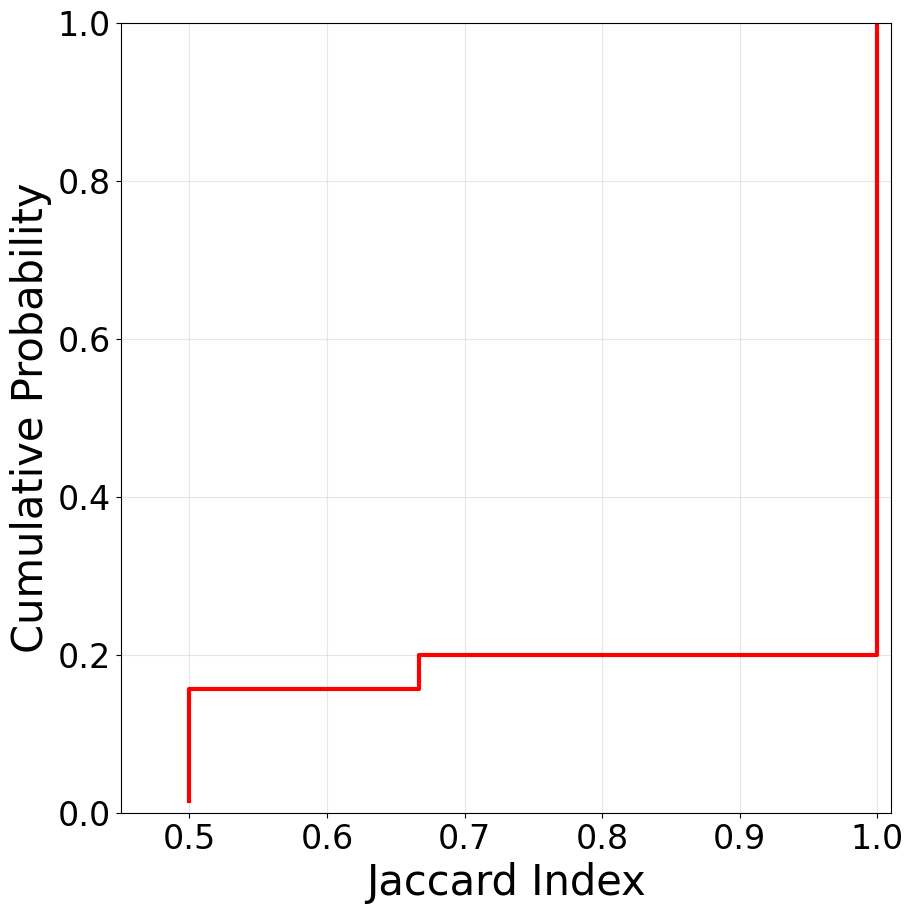}
	\end{minipage}
	\begin{minipage}[t]{0.28\linewidth}
		\centering
		\includegraphics[width=\textwidth]{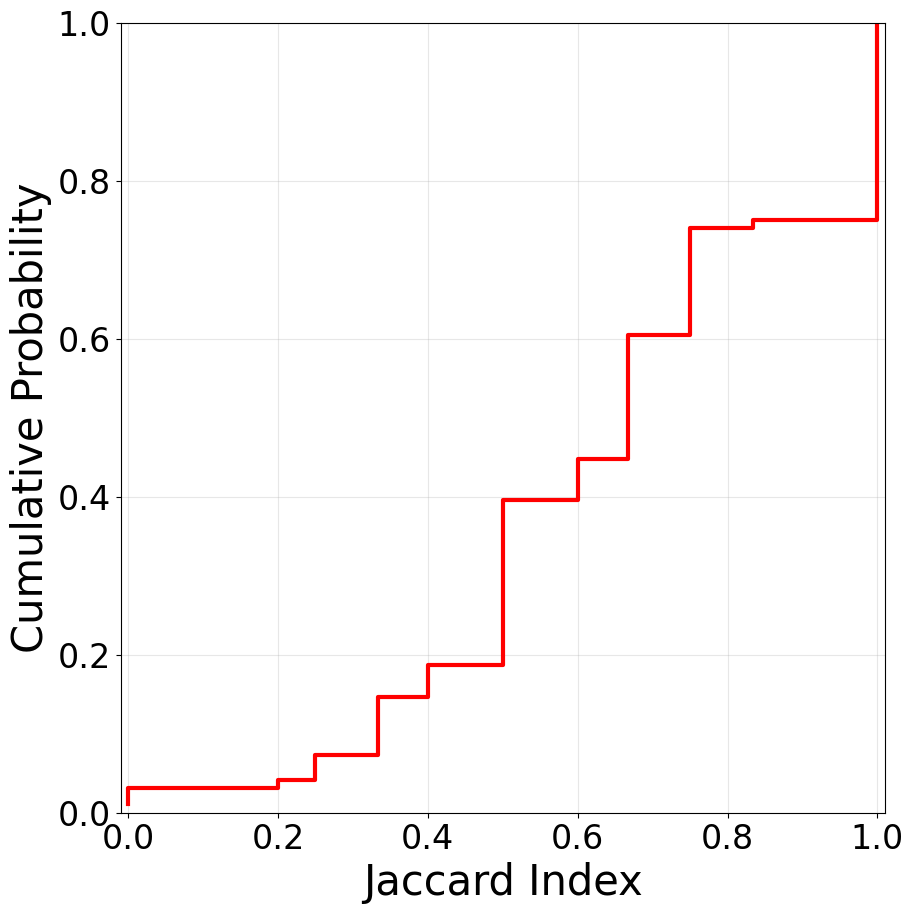}
	   \end{minipage}
	\caption{Cumulative distribution functions of Jaccard index scores measuring the overlap between relevant information source sets identified by our consensus annotations and those specified in the original dataset annotations. Jaccard indices were computed on the annotated subset for each dataset (HotPotQA left, MuSiQUE middle, MS-MARCO right). HotPotQA and MuSiQUE exhibit high agreement, with more than half the samples achieving perfect agreement. MS-MARCO shows 25\% are in perfect agreement.}
	\label{fig:agreement}
\end{figure}

Figure \ref{fig:agreement} shows the cumulative distribution functions of Jaccard index scores measuring agreement between our consensus annotations and the original dataset annotations for HotPotQA, MuSiQUE, and MS-MARCO. The Jaccard index quantifies the overlap between the sets of sources labeled as relevant. HotPotQA and MuSiQUE have binary annotations. For MS-MARCO, which uses a 0-3 relevance scale, we considered sources with scores of 2 or 3 as relevant. We had high agreement with annotations for HotPotQA and MuSiQUE, while MS-MARCO has moderate agreement.

\section{Ablations}
\label{app:ablations}

\begin{table*}[t]
	\centering
	\resizebox{\textwidth}{!}{%
	\begin{tabular}{|p{0.75\textwidth}|c|c|}
		\hline
		\textbf{Response} & \textbf{Mean Quality Score} & \textbf{Std. Dev} \\
		\hline
		{\footnotesize Based on the provided sources, I can confidently state: Connie May Fowler was \textcolor{red}{definitely} a memoirist. The source `Connie May Fowler' explicitly \textcolor{red}{states that she wrote memoirs, specifically mentioning} ``When Katie Wakes'' (which explores her family's generational cycle of domestic violence) and ``A Million Fragile Bones'' (about her life on a barrier island and the Deepwater Horizon oil spill).} & 0.3 & 0.0 \\
		\hline
		{\footnotesize Based on the provided sources, I can confidently state: Connie May Fowler was a memoirist. The source `Connie May Fowler' explicitly \textcolor{red}{describes her as a ``memoirist'' and mentions two of her memoirs}: ``When Katie Wakes'' (which explores her family's generational cycle of domestic violence) and ``A Million Fragile Bones'' (about her life on a barrier island and the Deepwater Horizon \textcolor{red}{BP} oil spill).} & 1.0 & 0.0 \\
		\hline
	\end{tabular}%
	}
	\caption{The LLM-as-a-judge \textsf{Quality} evaluation introduces sensitivity to token-level variations in semantically equivalent responses.
		Response 1 (top) was generated from four relevant sources. Response 2 (bottom) included one additional irrelevant source. Despite being semantically equivalent, the LLM-as-a-judge (Attribution LLM) assigned $\textsf{Quality}$ scores of 1.0 and 0.3 (scale: 0.0-1.0). The consistent scoring across 10 runs suggests that the LLMs are \emph{sensitive} to wording, but \emph{consistent} for the same wording.}
	\label{tab:inconsistency}
\end{table*}

\paragraph{Model Selection.}
\label{app:model_selection}

We evaluated three large language models for suitability, GPT-4.1 nano~\citep{openai2025gpt41}, Claude Haiku 3.5, and Claude Sonnet 4~\citep{anthropic2025claude4, anthropic2024haiku}, but conducted our main experiments using only the first two. As expected, attribution quality improved with model capability: Claude Haiku 3.5 achieved notably higher quality scores than GPT-4.1 nano at comparable token consumption levels across all Shapley algorithms (Figure \ref{fig:model_comparison}). However, the progression from Haiku 3.5 to Sonnet 4 deviated from this trend. While Sonnet 4 demonstrated greater token efficiency, it did not yield the anticipated improvement in attribution quality.

Investigation revealed that our prompts, optimized for GPT-4.1 nano and Haiku 3.5, proved overly restrictive for Sonnet 4. Specifically, instructions designed to prevent knowledge hallucination (e.g., directing the model not to fill knowledge gaps when sources cannot answer the query) were interpreted too strictly by Sonnet 4, causing it to refuse answering even when sources contained sufficient information. This suggests that prompt engineering requires model-specific calibration. More critically, Sonnet 4's higher cost--an order of magnitude greater than both GPT-4.1 nano and Haiku 3.5 (Figure \ref{fig:model_comparison})--combined with the extensive prompt re-engineering required, led us to exclude it from our main experiments.

To further assess model generalizability, we also evaluated two open-source Llama models on \name: a larger Llama 3.3-70B-Instruct-Turbo~\citep{grattafiori2024llama3} and a smaller Llama-3-8B-Instruct-Lite~\citep{grattafiori2024llama3}. The larger model achieved quality improvements comparable to those seen with stronger closed-source models, but at a cost on par with Sonnet 4, making it similarly impractical for our purposes. The smaller model underperformed even GPT-4.1 nano, suggesting that below a certain capability threshold, open-source models are not yet competitive for this task.

When we expanded our annotated dataset from 30 to 100 samples, we re-evaluated using GPT-4.1 nano and Claude Haiku 4.5~\citep{anthropic2024haiku}, as Haiku 3.5 had been retired in the interim. Haiku 4.5 outperforms its predecessor across all measured dimensions: despite carrying higher per-token pricing, it consumes fewer tokens overall, resulting in lower total cost, and is generally faster. In terms of attribution quality, Haiku 4.5 also improves over Haiku 3.5, placing it between GPT-4.1 nano and Haiku 3.5 in the overall model landscape (Figure \ref{fig:model_comparison}). Cost differences between GPT-4.1 nano and Haiku 4.5 remained modest, while GPT-4.1 nano retained its order-of-magnitude speed advantage per sample (Figure \ref{fig:model_comparison}). While API latency affects these measurements, the consistency of this difference suggests genuine efficiency advantages for time-sensitive applications.

\paragraph{Clipping.}
\label{app:clipping}
When comparing all attribution scores to ground truth relevance labels via Jaccard index, clipping has a minimal effect, with the largest difference being a 0.05 increase for FullShapley on HotPotQA with GPT-4.1 nano. However, clipping substantially improves Kendall $\tau_b$ ordinal correlation scores. Extremely small non-zero attribution scores (e.g., $<0.001$) introduce noise into ordinal correlation calculations by being treated as distinct ranked values rather than ties. Clipping eliminates this noise by setting near-zero attributions to exactly zero, resulting in clearer ordinal relationships. The most significant improvement was with MuSiQUE with Haiku 3.5, where the ordinal correlation between \name and FullShapley increased by 0.113 with clipping applied.

\paragraph{Caching.}
\label{app:caching}
We used caching in our baseline implementations to improve efficiency. For both FullShapley and the approximation baselines, we cached tested coalitions of sources and reused their LLM-as-a-judge scores upon cache hits to reduce costly LLM API calls. In FullShapley, caching was applied to sorted coalitions of sources--unlike the unsorted caching used in the approximation algorithms--which reduced redundant evaluations and required fewer coalition tests overall. This design choice improved cost and time efficiency: in an experiment with MuSiQUE, Haiku 3.5, and 10 data samples, unsorted caching resulted in a $3\times$ increase in token consumption, runtime, and therefore cost.

\begin{figure}[htbp]
	\centering
	\begin{minipage}{0.28\linewidth}
		\centering
		\textbf{MuSiQUE}
		\includegraphics[width=\textwidth]{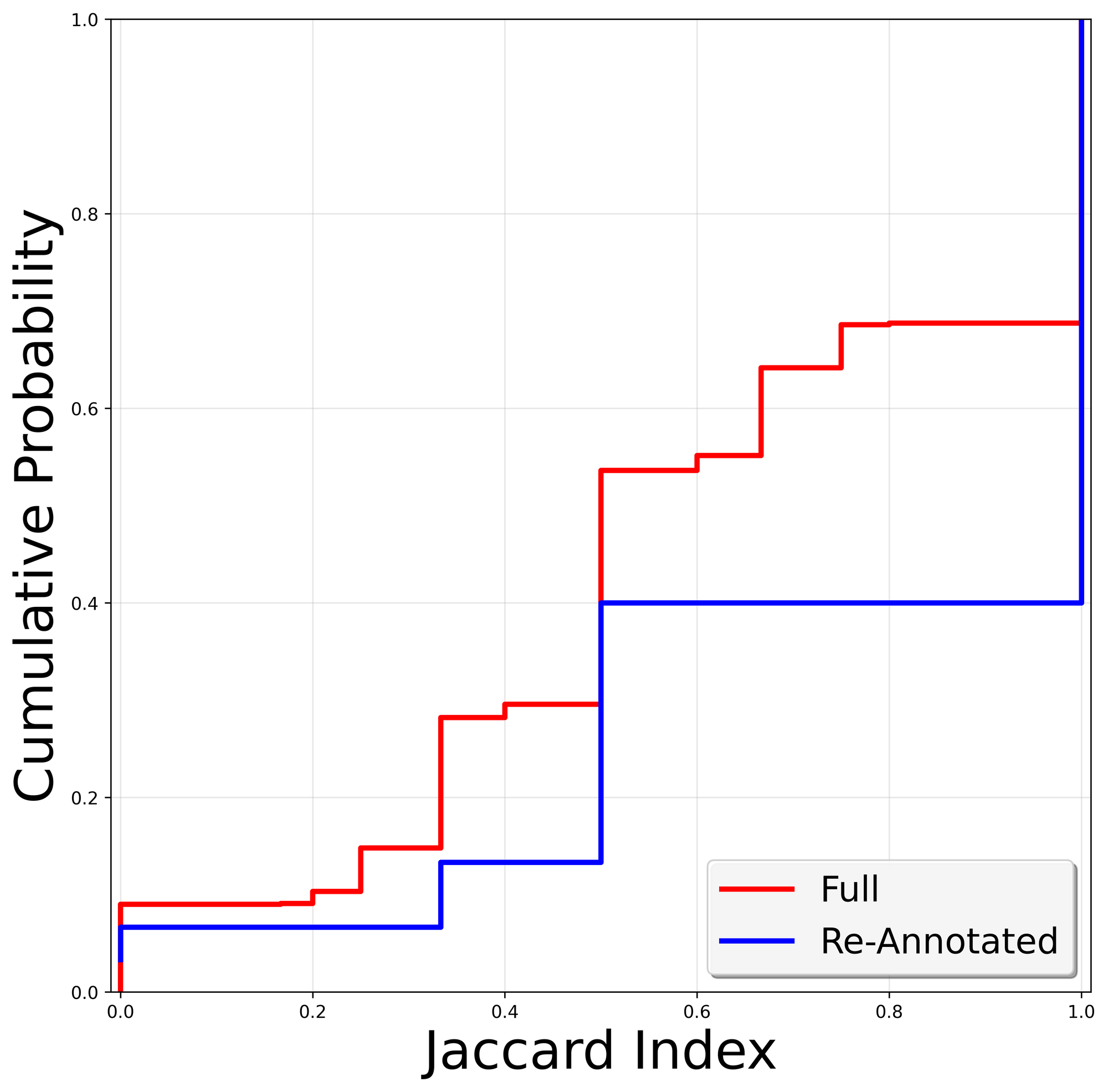}
	\end{minipage}
	\begin{minipage}{0.28\linewidth}
		\centering
		\textbf{HotPotQA}
		\includegraphics[width=\textwidth]{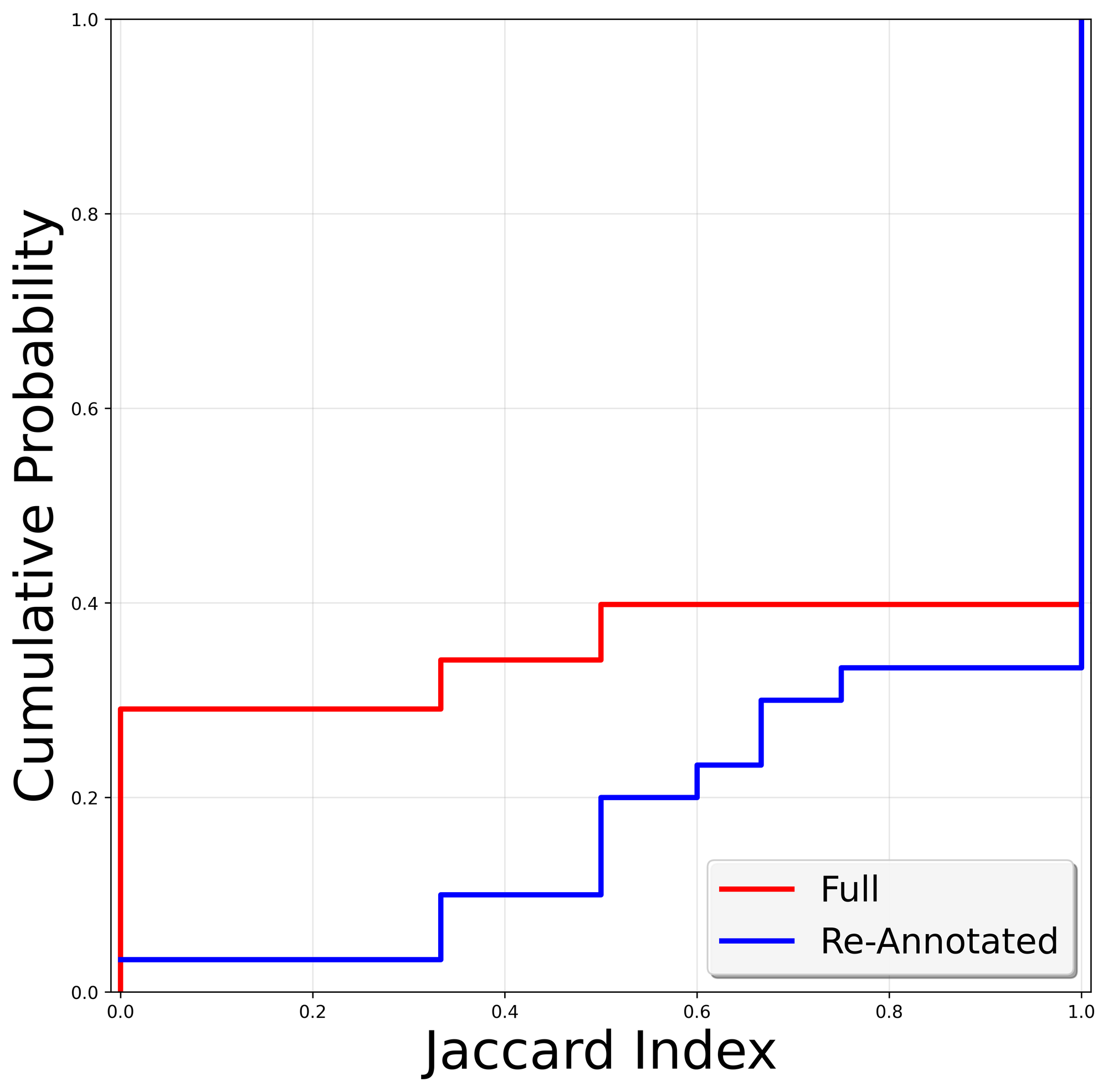}
		\end{minipage}
	\begin{minipage}{0.28\linewidth}
		\centering
		\textbf{MS-MARCO}
		\includegraphics[width=\textwidth]{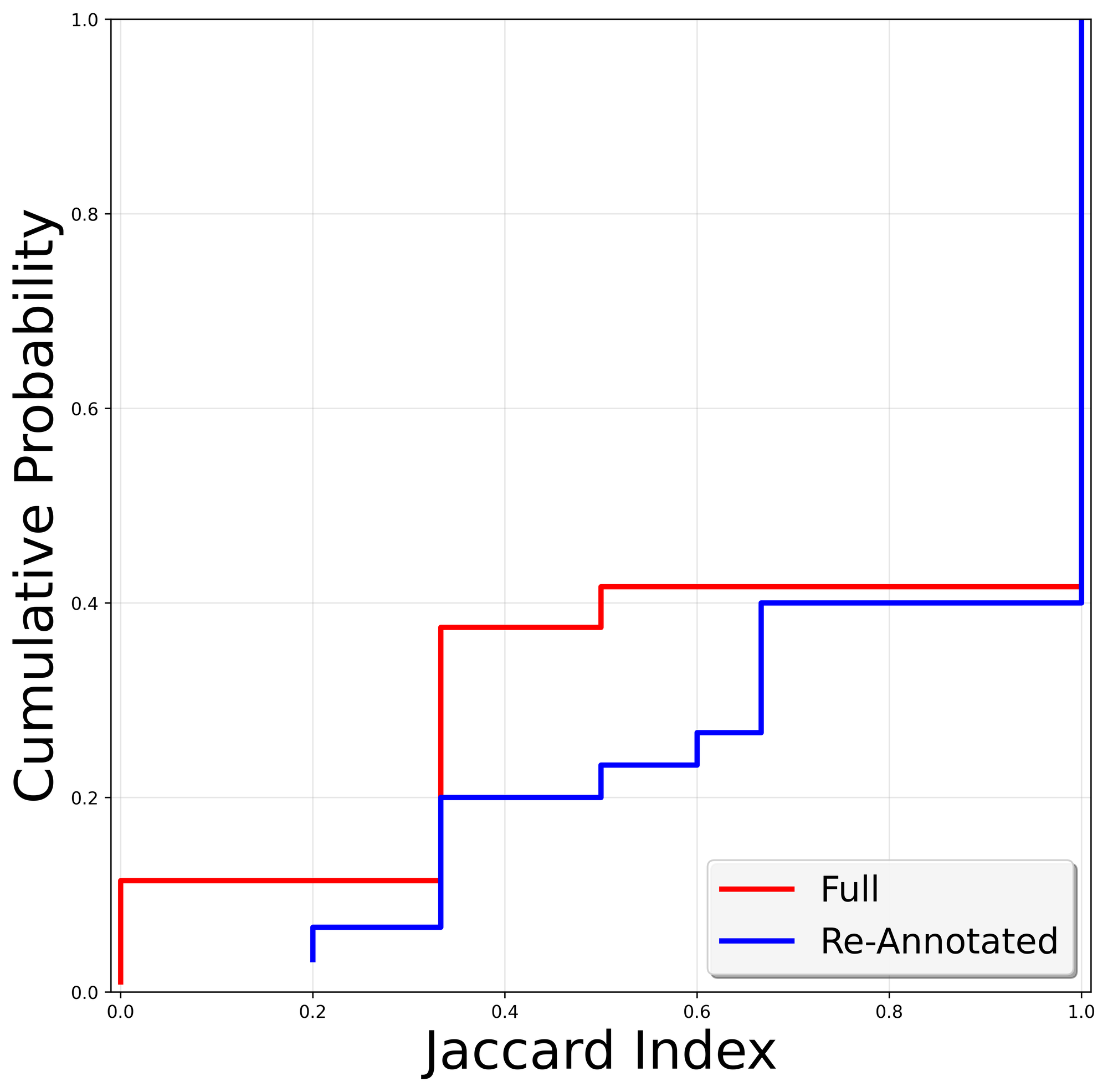}
	\end{minipage}
	\caption{Cumulative distribution function of Jaccard index scores between relevant information sources identified by \name and ground truth annotations from the full MuSiQUE answerable dataset (2,417 samples), HotPotQA dev dataset (7,405 samples), and MS-MARCO passages dataset with TREC 2019/2020 relevancy annotations (96 samples) with GPT-4.1 nano. The annotated data set results (on 30 data samples) are also depicted for comparison.
	}
	\label{fig:largeexp_cdf}
\end{figure}

\paragraph{Experiments on Large Datasets.}
\label{app:largeexp}
We conducted \name on the full MuSiQUE and HotPotQA dev datasets, and the MS-MARCO passages dataset with TREC 2019/2020 annotated datasets with GPT-4.1 nano, restricting our analysis to answerable queries (i.e., queries for which the provided information sources contain sufficient information to generate an answer).
Figure \ref{fig:largeexp_cdf} shows the cumulative distribution of Jaccard index scores across all 2,417 data samples for MuSiQUE, 7,405 data samples for HotPotQA, and a combined 96 data samples for MS-MARCO.
We observe a similar pattern to the agreement with our manually-annotated dataset, with more noise in the HotPotQA and MS-MARCO full datasets (this is expected, as we noted the original datasets often had noisy annotations, hence why we manually re-annotated a subset).
We observe a slightly noisier Jaccard index on the full MuSiQUE dataset, relative to our manually annotated subset. Although our manual annotations aligned completely with the original dataset labels, our annotated subset consisted primarily of 2-hop reasoning questions. When we evaluated the full MuSiQUE dataset, it also included 3-hop, 4-hop, and 5-hop questions, for which we observed a degradation in the average Jaccard index. This trend is consistent with prior observations that LLMs may exhibit reduced performance as the required reasoning depth increases ~\citep{lee2025gradegeneratingmultihopqa}, although our experiment does not isolate the specific source of this degradation. Nonetheless, the average Jaccard index for the full MuSiQUE development set remains $\ge 0.70$.

\begin{figure}[htbp]
	\centering
	\begin{minipage}{0.28\linewidth}
		\centering
		\textbf{MuSiQUE}
		\includegraphics[width=\textwidth]{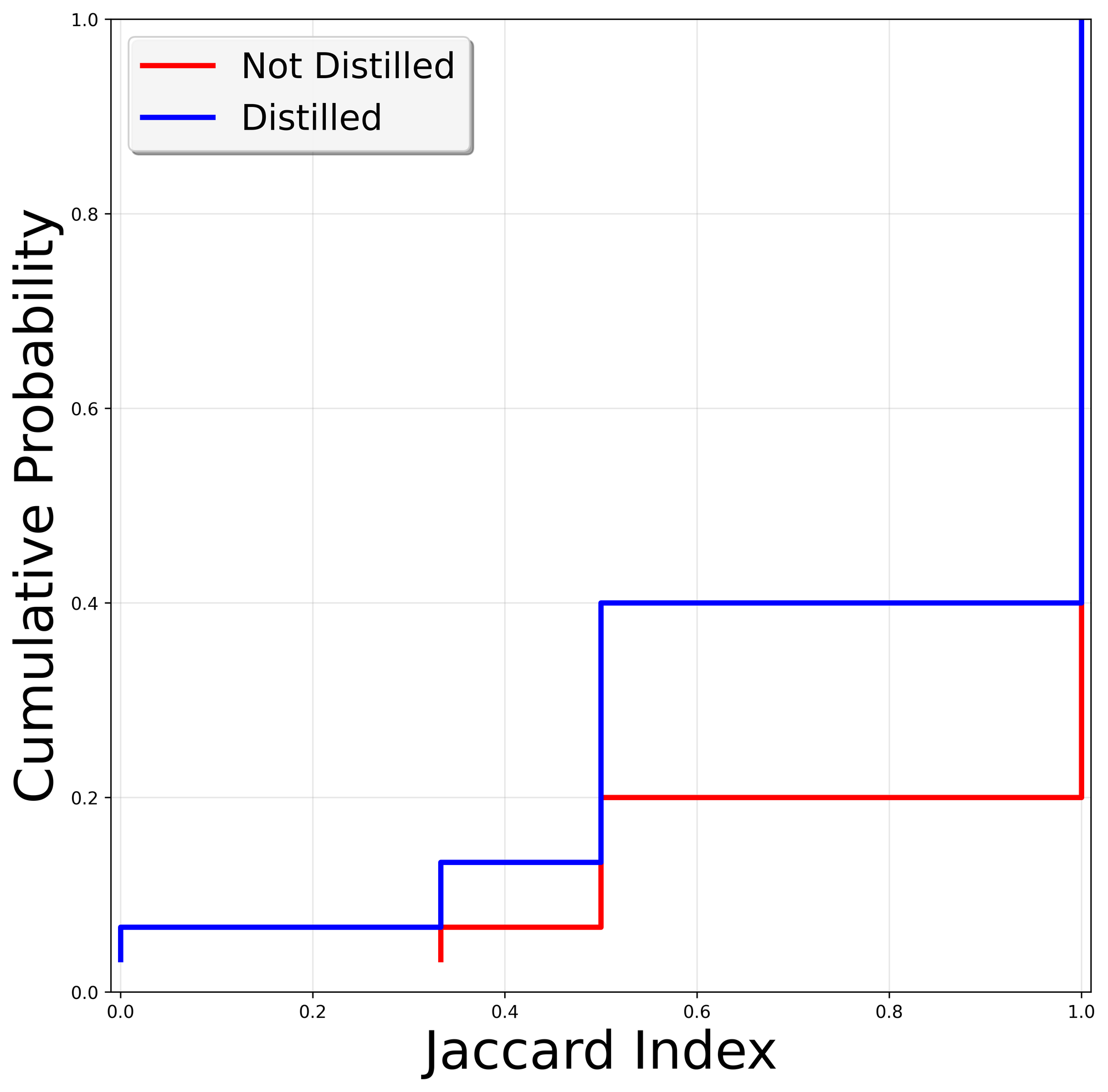}
	\end{minipage}
	\begin{minipage}{0.28\linewidth}
		\centering
		\textbf{HotPotQA}
		\includegraphics[width=\textwidth]{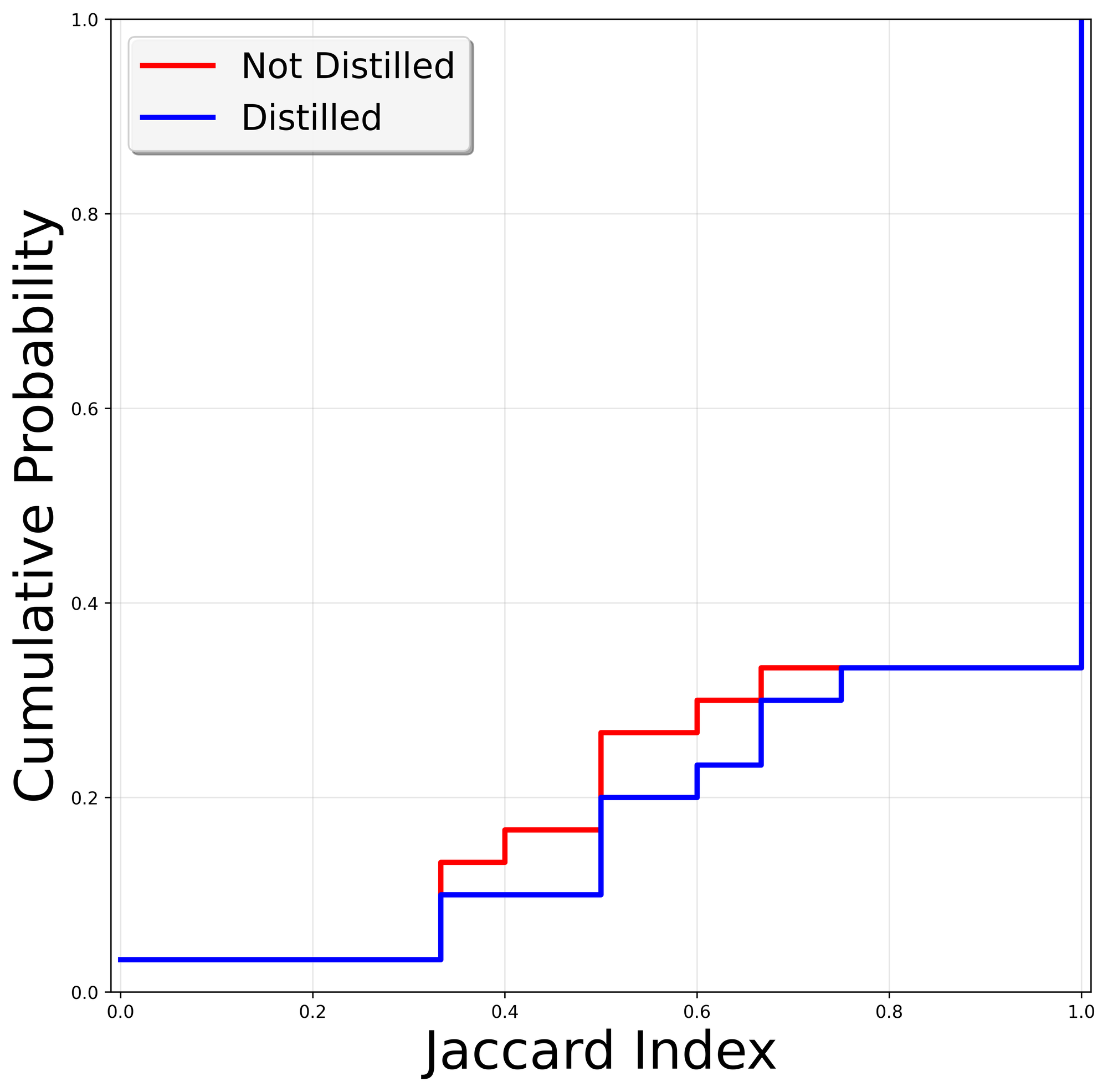}
		\end{minipage}
	\begin{minipage}{0.28\linewidth}
		\centering
		\textbf{MS-MARCO}
		\includegraphics[width=\textwidth]{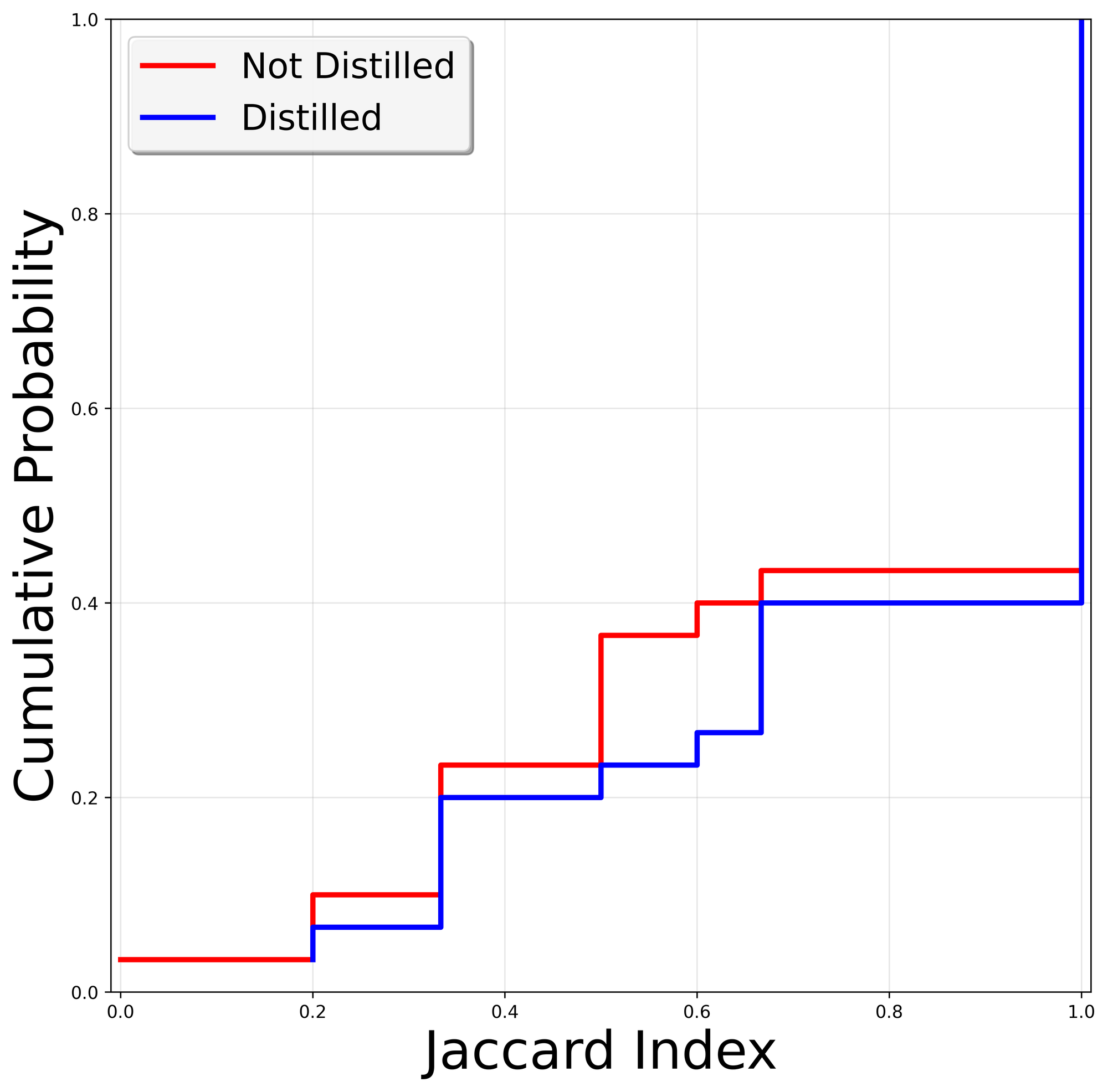}
	\end{minipage}
	\caption{Cumulative distribution function of Jaccard index scores between relevant information sources identified by \name and ground truth annotations between keypoint composition that are ``distilled" and ``not distilled" with GPT-4.1 nano. The average Jaccard index of MuSiQUE increases from 0.76 (distilled) to 0.89 (not distilled), HotPotQA declines from 0.83 to 0.81, and MS-MARCO declines from 0.78 to 0.73.
		}
	\label{fig:effectof_generalization}
\end{figure}

\paragraph{Impact of Keypoint Decomposition.}
\label{app:keypoint_decomp}
In our current implementation of keypoint decomposition, our prompt has a ``keypoint distillation" component, which filters out repetitive or redundant keypoints. To test the robustness of \name with different keypoint decomposition methodologies, we test \name on our 30-sample manually-annotated datasets with GPT-4.1 nano using the prompt from Figure \ref{fig:prompt2} without the  distillation component from Figure \ref{fig:prompt4}. The average Jaccard index changes by 0.02-0.13 across datasets. On MuSiQUE (Figure \ref{fig:effectof_generalization}), our results improve due to no distillation (0.13 increase). However, MS-MARCO and HotPotQA, which are more representative of ``messy" real-world web queries, suffer slightly (0.02-0.05 reduction) without distillation. This suggests that distillation is (slightly) helping the performance of \name. The robustness of \name in the face of different keypoint decomposition methodologies---including against adversarial manipulation---remains a direction for future research.

\begin{figure}[htbp]
	\centering
	\begin{minipage}[t]{0.7\linewidth}
	 \centering
	 \includegraphics[width=\textwidth]{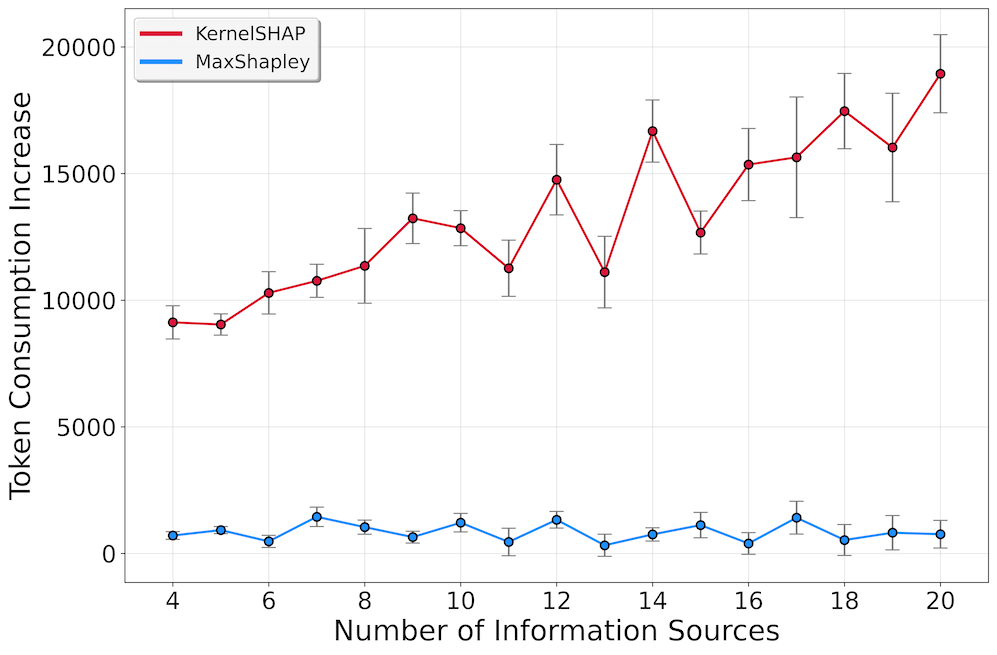}
	\end{minipage}
	\caption{
	Token consumption increase per additional information source for \name and KernelSHAP (calculated with 5 permutations of data), evaluated on MuSiQUE with GPT-4.1 nano. \name's marginal token consumption remains approximately constant as sources are added, indicating linear overall scaling. KernelSHAP's marginal cost grows with the number of sources, indicating super-linear scaling.
	}
	\label{fig:varying_sources}
\end{figure}

\paragraph{Scaling with Information Sources}
\label{app:scaling_exp}
A practical concern for \name is how computational cost scales with the number of information sources. We evaluate the scalability of \name against KernelSHAP (using 5 permutations of data) on MuSiQUE with GPT-4.1 nano, varying the number of sources (see Figure~\ref{fig:varying_sources}).

\name exhibits linear scaling in token consumption with respect to the number of sources, whereas KernelSHAP grows at a super-linear rate. While absolute costs may still be prohibitive for some production systems, understanding the scaling behavior of attribution methods is an important step toward assessing their feasibility in real-world deployments.

\paragraph{Additional Baselines: Sentence-Level Leave-One-Out}
\label{app:sentence_loo}
In our evaluation of \name, we use source-level Leave-One-Out (LOO) as a baseline, computing each source's contribution as $\phi^{\mathsf{LOO}}_i = U(S) - U(S \setminus \{s_i\})$. A natural variant is to compute attribution at the sentence level, removing individual sentences rather than entire sources: $\phi^{\mathsf{LOO}}_i = \frac{1}{|s_i|}\sum_{t_j \in s_i} \left[ U(S) - U(S \setminus \{t_j\}) \right]$. We evaluated this variant on HotPotQA with GPT-4.1 nano. Sentence-level LOO underperforms source-level LOO in attribution quality (Jaccard index of 0.26 vs.\ 0.52) while consuming significantly more tokens (~50k vs. ~15k). We therefore exclude it as a baseline in our main experiments.

\section{Implementation Considerations}
\label{app:implementation}

The actual implementation of Algorithm~\ref{alg:full-algorithm} can vary depending on the use scenario:
\begin{itemize}[leftmargin=*]
    \item \textbf{One-pass or Multiple Pass.} We can either ask the LLM to generate key points and scores in one pass within the same call during the answer generation process, or use multiple calls to the LLM to generate key points and scores separately.
    As in prior work, we used multiple calls to reduce hallucinations~\citep{gu2025surveyllmasajudge}.

    \item \textbf{Model Selection.} Given that the capability required for the LLM to generate key points and scores is weaker than the complete answer generation process, we can choose a fine-tuned LLM model or a smaller model for different purposes to further reduce computation cost.
    Our algorithm is designed to be agnostic to model selection; we show ablations in \Cref{app:ablations}.

    \item \textbf{Prompt Customization.} The prompts used in different stages can be customized to further improve performance under different use scenarios. In the generative search scenario, we can even adaptively generate score standards based on the user's query and retrieved sources to further improve score fidelity.
    We include the prompts for our implementation in \Cref{app:prompts}.
\end{itemize}

\section{Reproducibility}
\label{app:reproducibility}

To facilitate reproducibility and future research, we provide:
\begin{itemize}
	\item \textbf{Open-source code}: Full implementation of \name and all baselines, including evaluation scripts, all LLM prompts, and experiment configurations.\footnote{\url{https://github.com/spaddle-boat/MaxShapley}}
	\item \textbf{Educational demo}: An interactive web application demonstrating \name on sample queries.\footnote{\url{https://fair-search.com}}
	\item \textbf{Annotated datasets}: Manually re-annotated subsets of HotPotQA, MuSiQUE, and MS-MARCO (100 queries each) with consensus labels from independent annotators.
\end{itemize}


%



\end{document}